\documentclass{article}

\usepackage{amsmath}
\usepackage{microtype}
\usepackage{graphicx}
\usepackage{subfigure}
\usepackage{booktabs} % for professional tables

\usepackage{hyperref}

\usepackage{multirow}
\usepackage[table,xcdraw]{xcolor}
\usepackage[accepted]{icml2021}

\usepackage{amsmath}
\usepackage[normalem]{ulem}
\useunder{\uline}{\ul}{}

\icmltitlerunning{MFRS: A Multi-Frequency Reference Series Approach}

\begin{document}

\twocolumn[
\icmltitle{MFRS: A Multi-Frequency Reference Series Approach to Scalable and Accurate Time-Series Forecasting}



\icmlsetsymbol{equal}{*}

\begin{icmlauthorlist}
\icmlauthor{Liang Yu}{hust}
\icmlauthor{Lai Tu}{hust}
\icmlauthor{Xiang Bai}{hust}
% \icmlauthor{Iaesut Saoeu}{ed}
% \icmlauthor{Fiuea Rrrr}{to}
% \icmlauthor{Tateu H.~Yasehe}{ed,to,goo}
% \icmlauthor{Aaoeu Iasoh}{goo}
% \icmlauthor{Buiui Eueu}{ed}
% \icmlauthor{Aeuia Zzzz}{ed}
% \icmlauthor{Bieea C.~Yyyy}{to,goo}
% \icmlauthor{Teoau Xxxx}{ed}
% \icmlauthor{Eee Pppp}{ed}
\end{icmlauthorlist}

\icmlaffiliation{hust}{Huazhong University of Science and Technology}
% \icmlaffiliation{goo}{Googol ShallowMind, New London, Michigan, USA}
% \icmlaffiliation{ed}{School of Computation, University of Edenborrow, Edenborrow, United Kingdom}

% \icmlcorrespondingauthor{Cieua Vvvvv}{c.vvvvv@googol.com}
% \icmlcorrespondingauthor{Eee Pppp}{ep@eden.co.uk}

\icmlkeywords{Machine Learning, ICML}

\vskip 0.3in
]




\begin{abstract}
Multivariate time-series forecasting holds immense value across diverse applications, requiring methods to effectively capture complex temporal and inter-variable dynamics. 
A key challenge lies in uncovering the intrinsic patterns that govern predictability, beyond conventional designs, focusing on network architectures to explore latent relationships or temporal dependencies.
Inspired by signal decomposition, this paper posits that time series predictability is derived from periodic characteristics at different frequencies.
Consequently, we propose a novel time series forecasting method based on multi-frequency reference series correlation analysis. 
Through spectral analysis on long-term training data, we identify dominant spectral components and their harmonics to design base-pattern reference series.
Unlike signal decomposition, which represents the original series as a linear combination of basis signals, our method uses a transformer model to compute cross-attention between the original series and reference series, capturing essential features for forecasting.
Experiments on major open and synthetic datasets show state-of-the-art performance. 
Furthermore, by focusing on attention with a small number of reference series rather than pairwise variable attention, our method ensures scalability and broad applicability. 
The source code is available at: https://github.com/yuliang555/MFRS

\end{abstract}
\section{Introduction}
Time series forecasting plays a crucial role in a wide range of applications, from financial markets to weather prediction, energy demand forecasting, and beyond~\cite{angryk2020multivariate, chen2001freeway, khan2020towards}. Accurate forecasting is vital for making informed decisions in various sectors. The key challenge in time series forecasting lies in modeling the complex dependencies between time steps and, in the case of multivariate series, across different variables. Over the years, a variety of approaches have been proposed to tackle this challenge~\cite{Patchmixer, ModernTCN, PETformer, ETSformer, SOFTS}, many of which focus on designing advanced network architectures to capture these intricate relationships. Deep learning models, particularly those based on Recurrent Neural Networks (RNNs)~\cite{SegRNN, WITRAN} and Transformers~\cite{NSTransformer, TFT}, have achieved notable success by learning the temporal dynamics within data \cite{salinas2020deepar, LSTNet, Informer}. These models, while powerful, often concentrate on exploring the temporal dependencies between variables or uncovering complex inter-variable relationships to drive predictions.

Recent studies have also started to explore the frequency domain for time series forecasting, using techniques like Fourier Transform to extract periodic patterns from the data~\cite{FiLM, CoST}. These methods generally involve transforming the series into the frequency domain and using spectral features as additional dimensions for prediction \cite{FreTS, CoST}. However, while these techniques provide valuable insights into the periodic structure of time series data, they largely treat frequency components as standalone features, and still rely on traditional approaches for analyzing inter-variable or temporal relationships.

Inspired by Fourier Series, which represents periodic signals as a linear combination of sinusoidal components, 
%this paper takes a new perspective on time series forecasting. 
%Instead of focusing on pairwise variable dependencies or temporal relationships, 
we hypothesize that the key to accurate forecasting lies in identifying a set of common base patterns, which reflect the periodic behaviors across different variables, like the sinusoidal components for Fourier Series decomposition. 
By using the cross-attention mechanism of Transformer between the original time series and these base-pattern Reference Series~(RS), we characterize the relationship between the original series and the periodic patterns, thereby improving the forecasting accuracy.

By analyzing the spectrum of training data,
%and selecting key frequency components along with their harmonics
we can construct simple sinusoidal RS that serve as effective predictors. This approach, grounded in the idea that time series predictability stems from periodicity.
%, enables us to focus on the common underlying cyclical nature of the data. 
% While trend and seasonality are commonly regarded as the primary sources of predictability, our approach unifies them by treating trend as low-frequency components within a broader spectrum of periodicity. 
The resulting RS, derived from the dominant frequencies of a long training period, encapsulate the foundational periodicity that drives predictability in the data.

Building on this insight, we introduce the Multi-Frequency Reference Series (MFRS) method for time series forecasting. Our contributions are summarized as follows:
\begin{enumerate}
    \item We propose the MFRS method, which uses spectral analysis of long-term training data to extract the dominant frequency components of time series. These components, along with their harmonics, are used to design a set of base-pattern reference series. The Transformer model then computes the cross-attention between the original series and the base-pattern reference series, allowing us to capture the periodic nature of the signal for accurate forecasting.
    \item We introduce a base-pattern extraction algorithm that performs frequency selection based utilizing spectral analysis on the training time series. This algorithm is key in generating reference series that effectively capture the periodicity inherent in the data.
    \item We design a synchronization algorithm for the prediction phase, enabling fast alignment of the input series with reference series. This ensures accurate cross-attention calculation, even when predicting time series segments without explicit timestamps.
\end{enumerate}

Our method is simple yet effective, and we demonstrate its superior performance on several important open time series datasets as well as synthetic datasets, achieving state-of-the-art results. Notably, since the MFRS method computes attention between the original series and a small number of reference series, rather than calculating pairwise attention between multiple variables, it avoids the computational complexity that typically arises as the number of variables increases. This makes the method highly scalable and suitable for a wide range of time series forecasting applications.

\section{Related Work}

Various deep learning based methods have been proposed, leading to significant advancements in the field of time series forecasting, including RNN-based models \cite{salinas2020deepar, LSTNet, SutraNets, RWKV-TS}, CNN-based models \cite{liu2022scinet,wu2022timesnet,wang2023micn}, and MLP-based models \cite{TSMixer, huang2024hdmixer, TiDE, FITS}. Some studies also conduct research on the efficiency of pretrained Large Language Models (LLMs) in time series analysis tasks, such as LLMTime~\cite{LLMTime}, TEMPO~\cite{TEMPO}, OFA~\cite{OFA} and PromptCast~\cite{PromptCast}.

\textbf{Transformer-based Time Series Forecasting Models}~~~Recently, Transformer-based models have garnered extensive research due to the attention mechanism with powerful ability to capture dependency between pairs of elements. Early most of them focused on reducing the temporal and spatial complexity when constructing point-wise temporal dependency for long-term forecasting. Informer~\cite{Informer}, Autoformer~\cite{Autoformer}, Pyraformer~\cite{Pyraformer} and Triformer~\cite{Triformer} reduce the quadratic complexity to linear all by redesigning the attention mechanism. PatchTST~\cite{PatchTST}, on the other hand, utilizes patch-wise temporal dependency to improve the limitation of point-wise representations in capturing local information, and introduces the concept of channel independency for the first time. It is worth mentioning that, to respond the doubts of the validity of Transformer-based forecasters \cite{DLinear}, iTransformer~\cite{iTransformer} inverts the input structure of Transformer to capture channel dependency by taking the whole series as a token, for which the further receptive field and better portray of multivariate correlations are considered the key to its success.

\textbf{Methods of Utilizing Periodic Information}~~~Many studies have recognized the importance of periodic information in time series forecasting, of which the methods can be summarized into two categories: Seasonal-Trend Decomposition (STD) and Fourier Transform (FT) techniques. STD is used to effectively decouple the periodicity and trend from the original series. Classical approaches involve using kernel for moving aggregation, such as DLinear \cite{DLinear}, FEDformer \cite{Fedformer}, Autoformer \cite{Autoformer} with moving average (MOV) kernel and Leddam \cite{Leddam} with Learnable Decomposition (LD) kernel. Additionally, the recent Sparse technique in SparseTSF \cite{SparseTSF} and RCF in CycleNet \cite{CycleNet} are relatively novel types of STD that achieve impressive performance. There are various specific types of FT used to extract information in the frequency domain, such as Graph Fourier Transform (GFT) in StemGNN \cite{StemGNN}, Discrete Fourier Transform (DDT) in CoST \cite{CoST}, SFM~\cite{SFM} and FreTS \cite{FreTS}. Unlike the above methods, which embed the FT module into model for short-term operation, our proposed MFRS in this paper performs long-term Fast Fourier Transforms (FFT) before input, obtaining the inherent periodic patterns of series on a larger scale.

\begin{figure*}[htbp]
\centering
\centerline{\includegraphics[width=\textwidth]{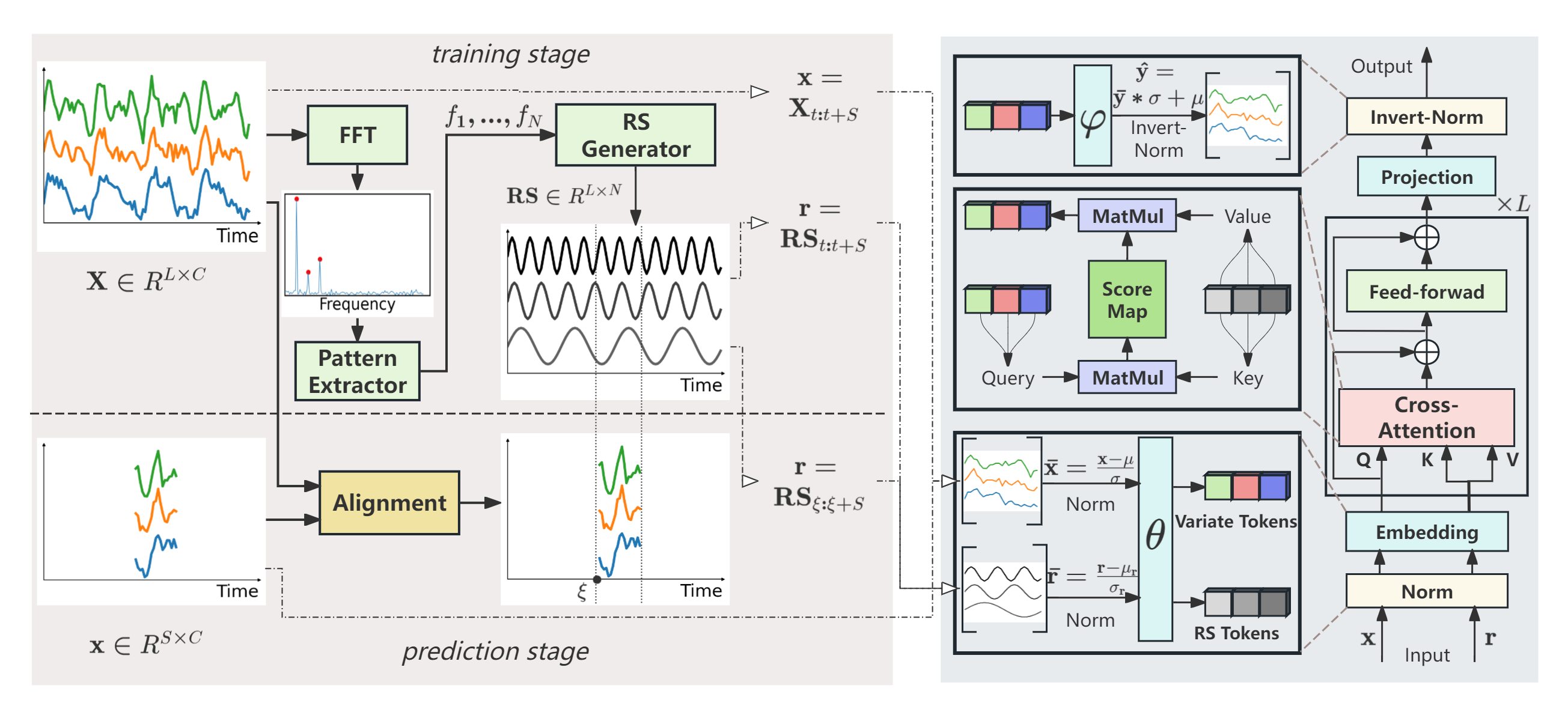}}
\caption{MFRS architecture. The left part shows the process of generating RS, which differs between training and prediction stage. The right part represents the modeling based on invert Transformer.}    
\label{figure-struc}
\vskip -0.05in
\end{figure*}

\section{Methodology}

A time series dataset with $C$ channels and $L$ steps is defined as $\mathbf{X} \in R^{L\times C}$. Given $S$ steps historical observations $\mathbf{X} _{t:t+S}=\left \{ \mathbf{X} _{t:t+S}^{\left (1 \right )},...,\mathbf{X} _{t:t+S}^{\left (C \right )}   \right \}\in R^{S\times C}$, the objective of multivariate time series forecasting is to predict $T$ future values $\mathbf{Y} _{t + S + 1:t +  S + T}\in R^{T\times C}$, which can both be abbreviated as lowercase $\mathbf{x}$ and $\mathbf{y}$ without subscripts. In this article, superscripts are uniformly used to represent one specific channel. For example, $\mathbf{X} _{t }^{\left ( i \right ) }$ denotes one $i$-th channel historical observation point at time $t$.

%%%%%%%%%%%%%%%%%%%%%%%%%%%%%%%%%%%%%%%%%%%%%%%%%%%%%%%%%%%%%%%%%%%%%%%%%%%%%%%
%%%%%%%%%%%%%%%%%%%%%%%%%%%%%%%%%%%%%%%%%%%%%%%%%%%%%%%%%%%%%%%%%%%%%%%%%%%%%%%
%                             Section 3.1
%%%%%%%%%%%%%%%%%%%%%%%%%%%%%%%%%%%%%%%%%%%%%%%%%%%%%%%%%%%%%%%%%%%%%%%%%%%%%%%
%%%%%%%%%%%%%%%%%%%%%%%%%%%%%%%%%%%%%%%%%%%%%%%%%%%%%%%%%%%%%%%%%%%%%%%%%%%%%%%
\subsection{Structure Overview}
\label{sec1-structure}

Accurately capturing the periodic information in time series will enhance the forecasting performance of model. The presence of periodicity in the time domain implies the existence of corresponding remarkable components in the frequency domain. A major highlight of this paper is the use of multi-frequency Reference Series containing the same periodic information to characterize series. MFRS achieves excellent performance by constructing correlation between the original series and RS, rather than relying on traditional temporal or channel dependency.

Specifically, the MFRS workflow is divided into two parts: obtaining RS as illustrated in the left of Figure~\ref{figure-struc} and modeling based on Transformer in the right. The method of obtaining RS differs between the training and prediction stages. During training, RS is generated from dataset $\mathbf{X}$ and sliced synchronously along with $\mathbf{X}$. The specific operation is accomplished by two well-designed modules: Base-Pattern Extractor~(\textbf{BPE}) and Reference-Series Generator~(\textbf{RSG}), of which the detail working principles will be introduced in Section~\ref{sec1-refs} later. In the prediction stage, historical data $\mathbf{x}$ should be aligned first with $\mathbf{X}$ to find the time step parameter $\xi$, which is essential to slice the already generated RS. The model in the right part takes both synchronized variables and RS as inputs, and explicitly extracts periodicity by establishing dependency among them through cross-attention. Here we adopt encoder-only Transformer, as a point has been argued that Transformer decoder may lead to performance degradation \cite{encoder-only}. Although the MFRS architecture does not undergo fundamental changes, new interpretations of some components are provided from a frequency domain perspective in Section~\ref{sec1-transformer} .

%%%%%%%%%%%%%%%%%%%%%%%%%%%%%%%%%%%%%%%%%%%%%%%%%%%%%%%%%%%%%%%%%%%%%%%%%%%%%%%
%%%%%%%%%%%%%%%%%%%%%%%%%%%%%%%%%%%%%%%%%%%%%%%%%%%%%%%%%%%%%%%%%%%%%%%%%%%%%%%
%                             Section 3.2
%%%%%%%%%%%%%%%%%%%%%%%%%%%%%%%%%%%%%%%%%%%%%%%%%%%%%%%%%%%%%%%%%%%%%%%%%%%%%%%
%%%%%%%%%%%%%%%%%%%%%%%%%%%%%%%%%%%%%%%%%%%%%%%%%%%%%%%%%%%%%%%%%%%%%%%%%%%%%%%

\subsection{Reference Series}
\label{sec1-refs}

\subsubsection{Spectrum Analysis}
\label{sec2-fft}
\begin{figure}[h]
\vskip -0.1in
\centering
\subfigure[]
{
    \begin{minipage}[b]{.47\linewidth}
        \label{figure-freq}
        \centering
        \includegraphics[width=\columnwidth]{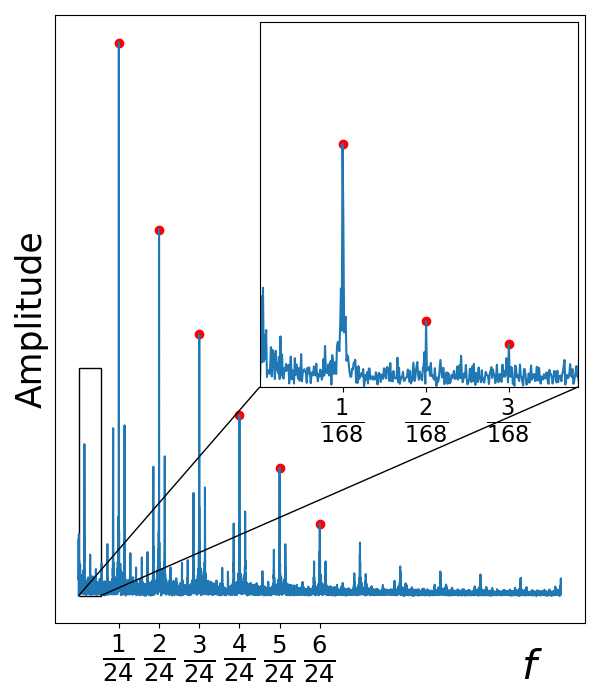}
    \end{minipage}
}
\subfigure[]
{
    \begin{minipage}[b]{.47\linewidth}
        \label{figure-cycle}
        \centering
        \includegraphics[width=\columnwidth]{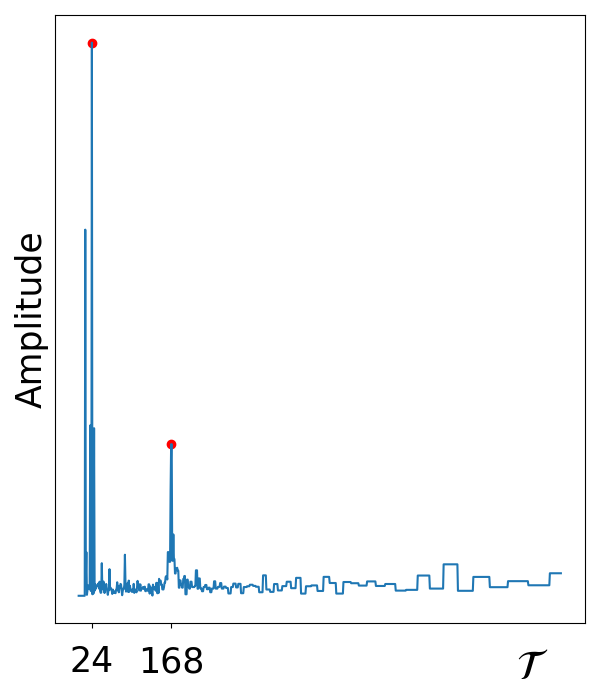}
    \end{minipage}
}
\vskip -0.1in
\caption{(a) spectrum $\Phi \left ( f  \right )$, used for extracting HBP.~~(b) converted spectrum $\Psi \left ( \mathcal{T}  \right )$, used for extracting PBP.}
\label{figure-fft}
\end{figure}

The obtained spectrum utilizing Fast Fourier Transform, should first be performed for module operation, as we are only concerned with the energy distribution of each component without considering the phase. Figure~\ref{figure-freq} shows the spectrum of one channel from \textbf{Traffic}, from which the signature can be summarized into two points: 1) The primary frequency component $f_{1}=\frac{1}{24}$ and $f_{2}=\frac{1}{168}$, indicate that the channel has periodic pattern daily and weekly, since time points are sampled hourly. This aligns with empirical knowledge that traffic flow tends to exhibit similar patterns each day and differs between weekdays and weekends. 2) Apart from $f_{1}$, $f_{2}$, some of their harmonic components, e.g, $2f_{1}$, $3f_{1}$, $4f_{1}$ and $2f_{2}$, $3f_{2}$, are also quite significant, which are caused by the step jump of the periodic sequence. Although not of any practical significance, they are also integral parts of the periodic patterns of the channel.

 Appendix~\ref{app1-spectrum} provides more detail spectrum analysis and shows that the periodic patterns among different channels are much of a muchness. It can be concluded that the periodic patterns within a time series consist of the primary base-patterns (PBP) that the period should be integers, and several harmonic base-patterns (HBP). Additionally, there are some relatively obvious sidelobes around base-patterns, which do not provide periodic information.

\subsubsection{Base-Pattern Extractor}
\label{sec2-bpe}

\textbf{Primary Base-Pattern}~~~The obtained spectrum of series with length $L$ can be expressed as a function: $\textrm{A}=\Phi \left ( f  \right ) ,f=\frac{l}{L},l\in \left ( 0,\frac{L}{2} \right ) \cap {Z} ^{+  }$, where $\textrm{A}$ denotes the amplitude. We convert it into a function taking the periodic pattern $\mathcal{T}$ as the variable, thus $\Psi \left ( \mathcal{T}  \right ) =\Phi \left ( \frac{1}{\mathcal{T}}  \right ),\mathcal{T}\in \left ( 0,L_{p} \right ) \cap {Z} ^{+  }$. Here, $L_{p}$ is a number much smaller than $L$. The conversion result of the spectrum in Figure~\ref{figure-freq} is shown in Figure~\ref{figure-cycle}.

The peaks in $\Psi$ include PBP, HBP and sidelobes, of which PBP has the highest amplitude. If $\Psi \left ( \mathcal{T}  \right )$ is largest among $\Psi \left ( 1  \right )\sim\Psi \left ( 2\mathcal{T}  \right )$, $\mathcal{T}$ is identified as a PBP. Then set all the values in this interval to zero to avoid interference with extracting the next PBP. Algorithm~\ref{alg-pbp} describes the process in detail.

\begin{algorithm}[h]
    \caption{Extracting PBP}
    \label{alg-pbp}
    \begin{algorithmic}
       \STATE {\bfseries Input:} spectrum $\Phi\left ( f  \right )$,
       \STATE ~~~~~~~~~~~~conversion length $L_{p}$, 
       \STATE {\bfseries Output:} PBP list $\Omega_{P}$
       \STATE Initialize $\Omega_{P}=\emptyset$
       \STATE $\Psi \left ( \mathcal{T}  \right ) =\Phi \left ( \frac{1}{\mathcal{T}}  \right ),\mathcal{T}=1,...,L _{p}$
       \FOR{$\mathcal{T}=2$ {\bfseries to} $\frac{L_{p}}{2}$}
       \IF{$\mathcal{T}=\arg \max\Psi \left [ :2\ast \mathcal{T}  \right ]$}
       \STATE insert $\mathcal{T}$ to $\Omega_{P}$
       \STATE Zero setting $\Psi \left [ :2\ast \mathcal{T}  \right ]$
       \ENDIF
       \ENDFOR
    \end{algorithmic}
\end{algorithm}

\textbf{Harmonic Base-Pattern}~~~Firstly we arrange the obtained PBP list $\Omega_{P}$ in ascending order as $f_{P}^{1}$,...,$f_{P}^{M}$. Then, the extraction of HBP employs a scoring mechanism. Let $\textrm{Score}\left ({kf_{P}^{m}} \right ), k=2,...K$ denotes the score for the $k$-th harmonic of PBP $f_{P}^{m}$. By traversing all channels, $\textrm{Score}\left ({kf_{P}^{m}} \right )$ is incremented by $\frac{\Phi\left ( kf_{P}^{m} \right ) }{\Phi\left ( f_{P}^{1} \right ) }$. It should be noted that when capturing HBP $f_{P}^{m+1}$, the impact of the sidelobes of $f_{P}^{m}$ needs to be removed, thus $K$ must satisfy $K*f_{P}^{m+1}<\frac{f_{P}^{m}}{2}$.  Finally, all extracted HBP are determined with top $Q$ hightest scores, where $Q$ is a manually set hyperparameter. The detail process is shown in Algorithm~\ref{alg-hbp}.

\begin{algorithm}[h]
    \caption{Extracting HBP}
    \label{alg-hbp}
    \begin{algorithmic}
       \STATE {\bfseries Input:} PBP list $\Omega_{P}$,~~channels C,~~hyperparameter $Q$
       \STATE {\bfseries Output:} HBP list $\Omega_{H}$
       \STATE Initialize $\Omega_{H}=\emptyset$ 
       \STATE arrange $\Omega_{P}$ in ascending order as $f_{P}^{1},...f_{P}^{M}$
       \FOR{$m=1$ {\bfseries to} $M$}
       \IF{m=1} 
       \STATE $K=\left \lfloor \frac{1}{2f_{P}^{1}}  \right \rfloor$
       \ELSE
       \STATE $K=\left \lfloor \frac{f_{P}^{m-1} }{2f_{P}^{m}}  \right \rfloor$
       \ENDIF
       \STATE Zero setting $\textrm{Score}\left ( kf_{P}^{m}  \right ),k=2,...K$
       \FOR{$c=1$ {\bfseries to} $C$}
       \STATE obtain spectrum $\Phi\left ( f  \right )$
       \FOR{$k=2$ {\bfseries to} $K$}
       \STATE $\textrm{Score}\left ( kf_{P}^{m}  \right ) =\textrm{Score}\left ( kf_{P}^{m}  \right )+ \frac{\Phi  \left ( {kf_{P}^{m}}  \right ) }{\Phi  \left ( {f_{P}^{1}}  \right ) }$
       \STATE insert $kf_{P}^{m}$ to $\Omega_{H}$
       \ENDFOR
       \ENDFOR
       \ENDFOR
       \STATE Sorting HBP in descending order based on scores
       \STATE $\Omega_{H}=\Omega_{H}[:Q]$
    \end{algorithmic}
\end{algorithm}

Additionally, there may be periodic patterns with long time spans in series, such as $\mathcal{T}=24\times 365=8760$ in \textbf{Traffic}. If the length of given series is insufficient, such patterns can be difficult to extract in spectrum. Therefore, we provide an option in \textbf{BPE} that allows us to set base-pattern manually based on empirical judgment.

\subsubsection{RS Generator}
\label{sec2-rsg}

All the extracted base-patterns are denoted as $\left \{ f_{1},...,f_{N}   \right \}$, including both PBP and HBP. Then \textbf{RSG} generates a set of single-cycle $\mathbf{RS} \in R^{L\times N}$, with length $L$ and frequencies ${f_{j} },j=1,...,N$. They can be in the form of sine, sawtooth, rectangle, or pulse, which essentially make no difference, as confirmed by the experimental results in Section~\ref{sec2-types}. We provide specific generation method for various types of RS in Appendix~\ref{app-refs}.

%%%%%%%%%%%%%%%%%%%%%%%%%%%%%%%%%%%%%%%%%%%%%%%%%%%%%%%%%%%%%%%%%%%%%%%%%%%%%%%
%%%%%%%%%%%%%%%%%%%%%%%%%%%%%%%%%%%%%%%%%%%%%%%%%%%%%%%%%%%%%%%%%%%%%%%%%%%%%%%
%                             Section 3.3
%%%%%%%%%%%%%%%%%%%%%%%%%%%%%%%%%%%%%%%%%%%%%%%%%%%%%%%%%%%%%%%%%%%%%%%%%%%%%%%
%%%%%%%%%%%%%%%%%%%%%%%%%%%%%%%%%%%%%%%%%%%%%%%%%%%%%%%%%%%%%%%%%%%%%%%%%%%%%%%
\subsection{Synchronous Alignment}
\label{sec1-syn}

The input of Transformer consists of two parts: historical observations $\mathbf{x}$ and RS $\mathbf{r}$. The complete information of signal is contained within three parameters: amplitude, frequency, and phase. For $\mathbf{r}$, the amplitude and frequency are fixed, while only the phase varies in accordance with the change in time step. This is the reason why $\mathbf{r}$ must be aligned with $\mathbf{x}$.

In the training phase, the alignment requirements are easily met. However, it may not necessarily be the case during the prediction, for which we are going to discuss. Generally speaking, the sampling time of $\mathbf{x}$ is known, and thus the time step can be obtained with respect to the first sampling point. In the contra case that the sampling time is indistinct, we utilize the signal processing synchronization technique to determine the time step parameter $\xi$.

\begin{figure}[h]
\begin{center}
\centerline{\includegraphics[width=0.5\textwidth]{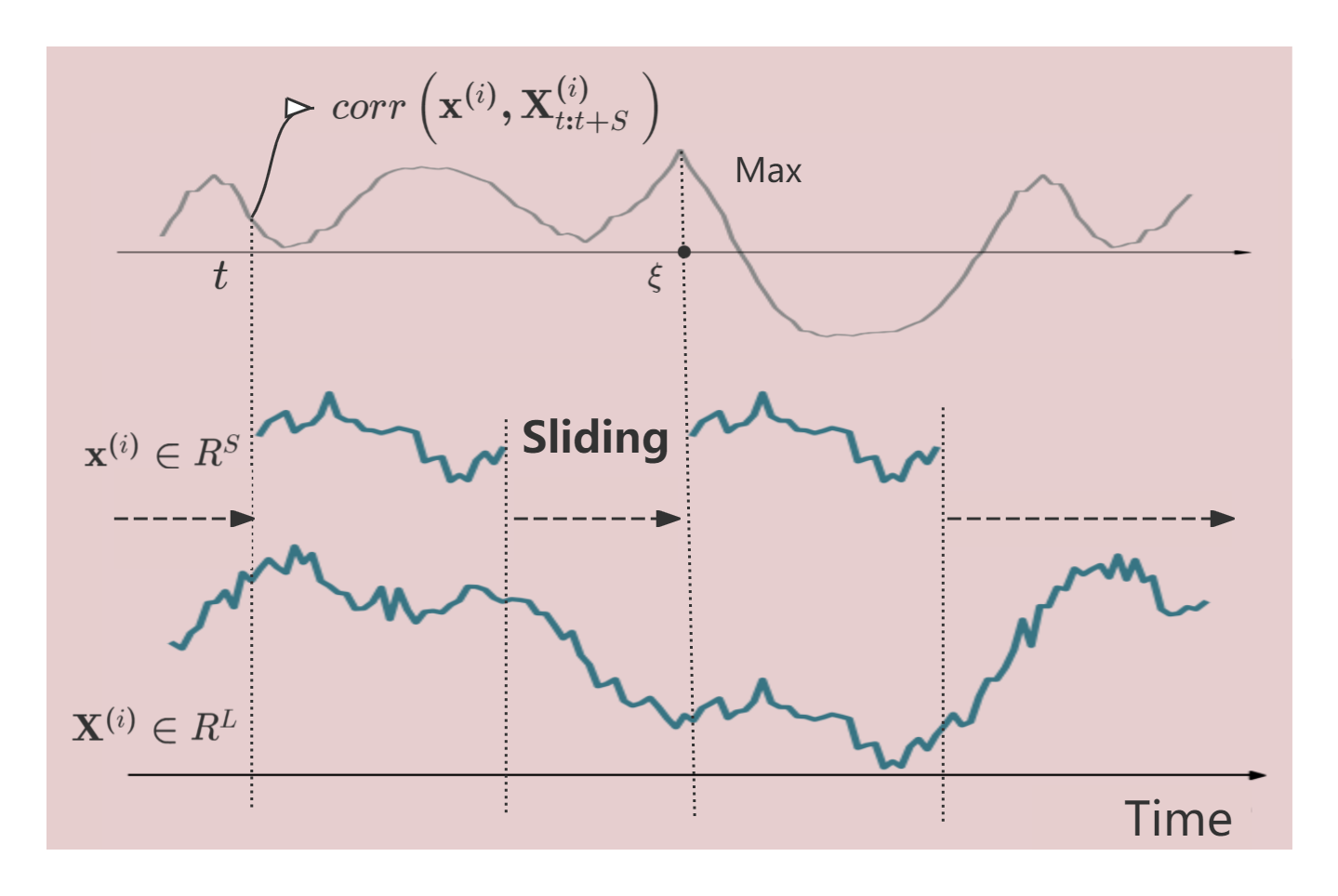}}
\caption{synchronous alignment method. sliding observation $\mathbf{x}^{\left ( i \right )}$ along with training data $\mathbf{X}^{\left ( i \right )}$ and calculating their similarity denoted as corr. Taking the time step with the largest corr as parameter $\xi$.}    
\label{figure-align}
\end{center}
\vskip -0.1in
\end{figure}

Intuitively, we believe that a univariate time series $\mathbf{X}^{\left ( i \right )}$ with periodicity exhibits a high similarity between two subsequences within the same time period. Consequently, we can slide and match an observation $\mathbf{x}^{\left ( i \right )}$ along $\mathbf{X}^{\left ( i \right )}$ to locate the segment with the highest similarity, thus determining the time step $\xi$. Figure~\ref{figure-align} visually demonstrates this sliding process, where we use the Pearson correlation coefficient to measure this similarity.

\begin{algorithm}[h]
    \caption{Alignment}
    \label{alg-syn}
    \begin{algorithmic}
       \STATE {\bfseries Input:} the max periodic pattern $\mathcal{T}_{M}$,
       \STATE ~~~~~~~~~~~~historical observations $\mathbf{x} \in R^{S\times C}$, \STATE ~~~~~~~~~~~~intercepted dataset $\mathbf{X}  \in R^{\left ( \mathcal{T}_{M}+  S   \right )\times C }$
       \STATE ~~~~~~~~~~~~channels $C$
       \STATE {\bfseries Output:} time step $\xi$
       \STATE Initialize $P_{t}=0,t=1,...,\mathcal{T} _{M}$.
       \FOR{$t=1$ {\bfseries to} $\mathcal{T} _{M}$}
       \FOR{$i=1$ {\bfseries to} $C$}
       \STATE $P_{t}=P_{t}+ corr\left ( \mathbf{x}^{\left ( i \right ) },\mathbf{X} _{t:t+S }^{\left ( i \right ) }  \right )$
       \ENDFOR
       \ENDFOR
       \STATE $\xi =\arg _t \max \left \{ P_{t}  \right \}$
    \end{algorithmic}
\end{algorithm}

Technically, for multivariate time series $\mathbf{X}$ and $\mathbf{x}$, one arbitrary channel is sufficient to get $\xi$ just following the above method, as the periodic patterns of all channels are similar. Without considering computational cost, multiple channels can be used to get more precise results. In this case, the time step $\xi$ is determined by finding the point with the maximal sum of correlation coefficients. Strictly speaking, the obtained parameter $\xi$ can only provide partial information about the sampling time of $\mathbf{x}$. Assuming that $\mathbf{x}$ exhibits a daily periodic pattern, we can determine its specific time within a day, but not a date. However, it makes no difference for our target $\mathbf{RS}_{\xi :\xi +S}$, of which each channel has a daily cycle. Let $\mathcal{T}_{max}=\max\left \{\mathcal{T}_{1},...,\mathcal{T}_{N}  \right \}$, then the parameter $\xi$ is given by Algorithm~\ref{alg-syn}.

%%%%%%%%%%%%%%%%%%%%%%%%%%%%%%%%%%%%%%%%%%%%%%%%%%%%%%%%%%%%%%%%%%%%%%%%%%%%%%%
%%%%%%%%%%%%%%%%%%%%%%%%%%%%%%%%%%%%%%%%%%%%%%%%%%%%%%%%%%%%%%%%%%%%%%%%%%%%%%%
%                             Section 3.4
%%%%%%%%%%%%%%%%%%%%%%%%%%%%%%%%%%%%%%%%%%%%%%%%%%%%%%%%%%%%%%%%%%%%%%%%%%%%%%%
%%%%%%%%%%%%%%%%%%%%%%%%%%%%%%%%%%%%%%%%%%%%%%%%%%%%%%%%%%%%%%%%%%%%%%%%%%%%%%%
\subsection{Transformer}
\label{sec1-transformer}

\textbf{Norm \& Embedding}~~~~The correlation between variates and RS arises from their shared frequency component. Shot-term signals generally contain abundant DC components, which can introduce noise into the modeling process. Therefore, all signals need to undergo DC-blocking filtration before embedding. Therefore, the first step in modeling is to perform the LayerNorm operation, which can also be considered as a kind of filtering method.

MFRS adopts an invert-transformer architecture, taking the whole series as a token. Embedding: $R^{S} \longrightarrow R^{D}$ maps tokens to the hidden space of encoder, where $D$ denotes the dimension of hidden space. Alternatively, there could be another interpretation that embedding transforms time-domain signals containing position and numerical information into frequency-domain signals that incorporate amplitude, frequency, and phase information. Moreover, numerous experiments have confirmed that whether variate and RS share an Embedding layer or not has little impact on model performance.

\textbf{Cross Attention}~~~~In time series forecasting, transformer is used to establish temporal and channel dependency mainly based on self-attention. MFRS leverages cross-attention to model the nonlinear correlations hidden behind variates and RS. Specifically, variates are transformed into queries $\mathbf{Q} \in R^{C\times D}$ through linear projection, while RS into keys and values $\mathbf{K,V} \in R^{N\times D}$. The score map obtained can be described as $\mathbf{A}=\left ( \mathbf{Q} \mathbf{K} ^{T}/\sqrt{D}   \right ) \in R^{C\times N}$. Then we acquire tokens representing variates through the interaction between $\mathbf{A}$ and values $\mathbf{V}$.

\textbf{Projection \& Invert-Norm}~~~~The TrmBlocks output the frequency-domain representation of predicted tokens, then Projection: $R^{D} \longrightarrow R^{T}$ maps them back to the original space. Finally, the DC components are injected back into tokens through the Invert-Norm operation, which can be expressed as

\begin{equation*}
    \centering
    \hat{\mathbf{y}}^{\left ( i \right ) }=\bar{\mathbf{y}} ^{\left ( i \right ) }\ast \sqrt{\textrm{Var}\left ( \mathbf{x} ^{\left ( i \right ) }  \right ) }+  \textrm{Mean}\left ( \mathbf{x} ^{\left ( i \right ) } \right )
\end{equation*}

\section{Experiments}

%%%%%%%%%%%%%%%%%%%%%%%%%%%%%%%%%%%%%%%%%%%%%%%%%%%%%%%%%%%%%%%%%%%%%%%%%%%%%%%
%%%%%%%%%%%%%%%%%%%%%%%%%%%%%%%%%%%%%%%%%%%%%%%%%%%%%%%%%%%%%%%%%%%%%%%%%%%%%%%
%                             datasets
%%%%%%%%%%%%%%%%%%%%%%%%%%%%%%%%%%%%%%%%%%%%%%%%%%%%%%%%%%%%%%%%%%%%%%%%%%%%%%%
%%%%%%%%%%%%%%%%%%%%%%%%%%%%%%%%%%%%%%%%%%%%%%%%%%%%%%%%%%%%%%%%%%%%%%%%%%%%%%%
\subsection*{Datasets}
\label{sec1-datasets}

We gather eight open datasets widely used in time series forecasting research, including 4 ETT datasets (\textbf{ETTh1}, \textbf{ETTh2}, \textbf{ETTm1}, \textbf{ETTm2}), \textbf{Electricity}, \textbf{Weather}, \textbf{Traffic}, \textbf{Solar}, which cover various areas in real world. we validate the performance of our proposed MFRS on them, of which the detailed information is exhibited in Appendix~\ref{app1-description}

To further compare performance among various forecasters, we additionally constructed several synthetic datasets \textbf{Compose}, where signal $\mathbf{X}$ consists of two components: deterministic signal $\mathbf{Z }$ with a definite value at any given time and random signal $\mathbf{U}$ whose distribution does not change over time (the trend component will be left for future studies). Specifically, $\mathbf{Z }$ is synthesized by summing four sinusoidal sequences with periods of $\mathcal{T} \in \left \{ 72, 36, 24, 18 \right \}$ or $\left \{ 720, 360, 240, 180 \right \}$, and $\mathbf{U} \sim N\left ( 0,\sigma ^{2}  \right )$ follows a Gaussian distribution. Unlike open datasets, the parameters of the random component in \textbf{Compose} are known, and a set of different datasets can be obtained by varying the parameter $\sigma$. Appendix~\ref{app2-syn-method} provides a specific method of generating more synthetic datasets.

%%%%%%%%%%%%%%%%%%%%%%%%%%%%%%%%%%%%%%%%%%%%%%%%%%%%%%%%%%%%%%%%%%%%%%%%%%%%%%%
%%%%%%%%%%%%%%%%%%%%%%%%%%%%%%%%%%%%%%%%%%%%%%%%%%%%%%%%%%%%%%%%%%%%%%%%%%%%%%%
%                             Section 4.1
%%%%%%%%%%%%%%%%%%%%%%%%%%%%%%%%%%%%%%%%%%%%%%%%%%%%%%%%%%%%%%%%%%%%%%%%%%%%%%%
%%%%%%%%%%%%%%%%%%%%%%%%%%%%%%%%%%%%%%%%%%%%%%%%%%%%%%%%%%%%%%%%%%%%%%%%%%%%%%%
\subsection{Results of Open Datasets}
\label{sec1-open}

% \usepackage[normalem]{ulem}

\begin{table*}[h]\small
\setlength{\tabcolsep}{3pt}
\renewcommand{\arraystretch}{1.5}
\centering
\caption{Multivariate time series forecasting results on eight open datasets with fixed look-back length $S=96$ and prediction horizons $T\in \left \{ 96, 192, 336, 720 \right \}$. Results are averaged from all prediction horizons}
\vskip 0.15in
\label{table_all_avg}
\begin{tabular}{ccccccccccccccccc}
\hline
datasets & \multicolumn{2}{c}{\textbf{ETTm1}} & \multicolumn{2}{c}{\textbf{ETTm2}} & \multicolumn{2}{c}{\textbf{ETTh1}} & \multicolumn{2}{c}{\textbf{ETTh2}} & \multicolumn{2}{c}{\textbf{Electricity}} & \multicolumn{2}{c}{\textbf{Traffic}} & \multicolumn{2}{c}{\textbf{Weather}} & \multicolumn{2}{c}{\textbf{Solar}} \\
metrics & \textbf{MSE} & \textbf{MAE} & \textbf{MSE} & \textbf{MAE} & \textbf{MSE} & \textbf{MAE} & \textbf{MSE} & \textbf{MAE} & \textbf{MSE} & \textbf{MAE} & \textbf{MSE} & \textbf{MAE} & \textbf{MSE} & \textbf{MAE} & \textbf{MSE} & \textbf{MAE} \\ \hline
MFRS(ours) & \textbf{0.372} & \textbf{0.39} & {\ul 0.276} & {\ul 0.323} & \textbf{0.429} & \textbf{0.431} & \textbf{0.369} & \textbf{0.396} & \textbf{0.161} & \textbf{0.256} & \textbf{0.409} & \textbf{0.27} & \textbf{0.242} & \textbf{0.269} & {\ul 0.225} & \textbf{0.257} \\ \hline
CycleNet/MLP & {\ul 0.379} & {\ul 0.396} & \textbf{0.266} & \textbf{0.314} & 0.457 & {\ul 0.44} & 0.388 & 0.409 & {\ul 0.168} & {\ul 0.259} & 0.472 & 0.301 & {\ul 0.243} & {\ul 0.271} & \textbf{0.21} & {\ul 0.261} \\
iTransformer & 0.407 & 0.41 & 0.288 & 0.332 & 0.454 & 0.448 & {\ul 0.383} & {\ul 0.406} & 0.178 & 0.27 & {\ul 0.428} & {\ul 0.282} & 0.258 & 0.278 & 0.233 & 0.262 \\
PatchTST & 0.387 & 0.4 & 0.281 & 0.326 & 0.469 & 0.454 & 0.387 & 0.407 & 0.205 & 0.29 & 0.481 & 0.304 & 0.258 & 0.28 & 0.27 & 0.307 \\
DLinear & 0.403 & 0.407 & 0.35 & 0.401 & 0.456 & 0.452 & 0.559 & 0.515 & 0.212 & 0.3 & 0.624 & 0.383 & 0.265 & 0.317 & 0.33 & 0.401 \\
FEDformer & 0.448 & 0.452 & 0.304 & 0.349 & {\ul 0.44} & 0.46 & 0.436 & 0.449 & 0.214 & 0.327 & 0.609 & 0.376 & 0.309 & 0.36 & 0.292 & 0.381 \\
TimesNet & 0.4 & 0.406 & 0.291 & 0.332 & 0.458 & 0.45 & 0.414 & 0.427 & 0.192 & 0.295 & 0.62 & 0.336 & 0.259 & 0.286 & 0.301 & 0.319 \\
TiDE & 0.419 & 0.419 & 0.358 & 0.404 & 0.541 & 0.507 & 0.611 & 0.554 & 0.252 & 0.344 & 0.76 & 0.473 & 0.27 & 0.32 & 0.347 & 0.418 \\
SCINet & 0.486 & 0.481 & 0.57 & 0.537 & 0.747 & 0.647 & 0.954 & 0.723 & 0.268 & 0.365 & 0.804 & 0.509 & 0.292 & 0.363 & 0.282 & 0.375 \\
Crossformer & 0.513 & 0.495 & 0.757 & 0.61 & 0.529 & 0.522 & 0.942 & 0.684 & 0.244 & 0.334 & 0.55 & 0.304 & 0.258 & 0.315 & 0.641 & 0.639 \\
Autoformer & 0.588 & 0.517 & 0.327 & 0.371 & 0.496 & 0.487 & 0.45 & 0.459 & 0.227 & 0.338 & 0.628 & 0.379 & 0.338 & 0.382 & 0.885 & 0.711 \\ \hline
\end{tabular}
\end{table*}

\textbf{Baselines \& Settings}~~~We select five Transformer-Based models, including iTransformer, PatchTST, Crossformer, Autoformer and FEDformer, two Linear-Based models CycleNet, DLinear and TiDE, two TCN-Based models SCINet and TimesNet, totaling ten SOTA forecasters as our baselines. All models follow the same experimental setup with a fixed look-back window $S=96$ and prediction lengths $T\in \left \{ 96,192,336,720 \right \}$. We will use the calculated MSE and MAE as evaluation metrics, where lower values indicate better model performance.

\textbf{Results}~~~Table~\ref{table_all_avg} shows the detailed experimental results of all models. The best results are highlighted in \textbf{bold} and the second best are \uline{underline}. We can observe that MFRS exhibits exceptional performance across all datasets, demonstrating its robust ability to capture periodic information. Specifically, MFRS only slightly lags behind CycleNet in performance on \textbf{ETTm2} and \textbf{Solar}, while achieving the best results on the remaining datasets. Especially on \textbf{Traffic}, which has a strong periodicity, MFRS achieves significant improvements of 0.019 in MSE and 0.012 in MAE respectively, compared to the second best model. Additionally, we notice that CycleNet, which focuses on capturing periodic information, also yields impressive prediction results, but still has a certain gap compared to MFRS. This further highlights the efficiency and excellent stability of MFRS.

%%%%%%%%%%%%%%%%%%%%%%%%%%%%%%%%%%%%%%%%%%%%%%%%%%%%%%%%%%%%%%%%%%%%%%%%%%%%%%%
%%%%%%%%%%%%%%%%%%%%%%%%%%%%%%%%%%%%%%%%%%%%%%%%%%%%%%%%%%%%%%%%%%%%%%%%%%%%%%%
%                             Section 4.2
%%%%%%%%%%%%%%%%%%%%%%%%%%%%%%%%%%%%%%%%%%%%%%%%%%%%%%%%%%%%%%%%%%%%%%%%%%%%%%%
%%%%%%%%%%%%%%%%%%%%%%%%%%%%%%%%%%%%%%%%%%%%%%%%%%%%%%%%%%%%%%%%%%%%%%%%%%%%%%%
\subsection{Results of Synthetic Datasets}
\label{sec1-synthetic}

Compared to open datasets, our synthetic ones offer the advantage of being able to derive theoretical optimal predictions, which can serve as a novel benchmark. Let $\mathbf{E} \left ( \cdot  \right )$ denotes the expectation of random variable. Among two parts of signal $\mathbf{X }$, $\mathbf{Z }$ is completely predictable, meaning that the optimal prediction value $\hat{\mathbf{Z} }_{t} =\mathbf{Z}_{t}$. On the contrary, $\mathbf{U}$ is completely unpredictable, and the optimal outcome $\hat{\mathbf{U}}=\mathbf{E} \left ( \mathbf{U}  \right )$ as demonstrated in the Appendix~\ref{app2-random}. Based on the above assertion, we can derive the theoretically optimal evaluation metric $\textrm{MSE}_\textrm{Optimal}=\sigma ^{2}$ and $\textrm{MAE}_\textrm{Optimal}=\sqrt{\frac{2}{\pi } }\sigma$. The derivation process can be referred to in Appendix~\ref{app2-optimal}.

This section presents the experimental results visually for \textbf{Compose}. By altering the parameter of Gaussian distribution as $\sigma\in \left \{ 0,1,2,3,4,5 \right \}$, multiple distinct \textbf{Compose} can be obtained, each of which is then experimented on MFRS, iTransformer and DLinear. The three models follow the same experimental setup with a fixed look-back window $S=96$ and prediction lengths $T\in \left \{ 96,720 \right \}$.

\begin{figure}[h]
\vskip -0.1in
\centering
\subfigure[T~=~96]
{
    \begin{minipage}[b]{0.47\linewidth}
        \centering
        \includegraphics[width=\columnwidth]{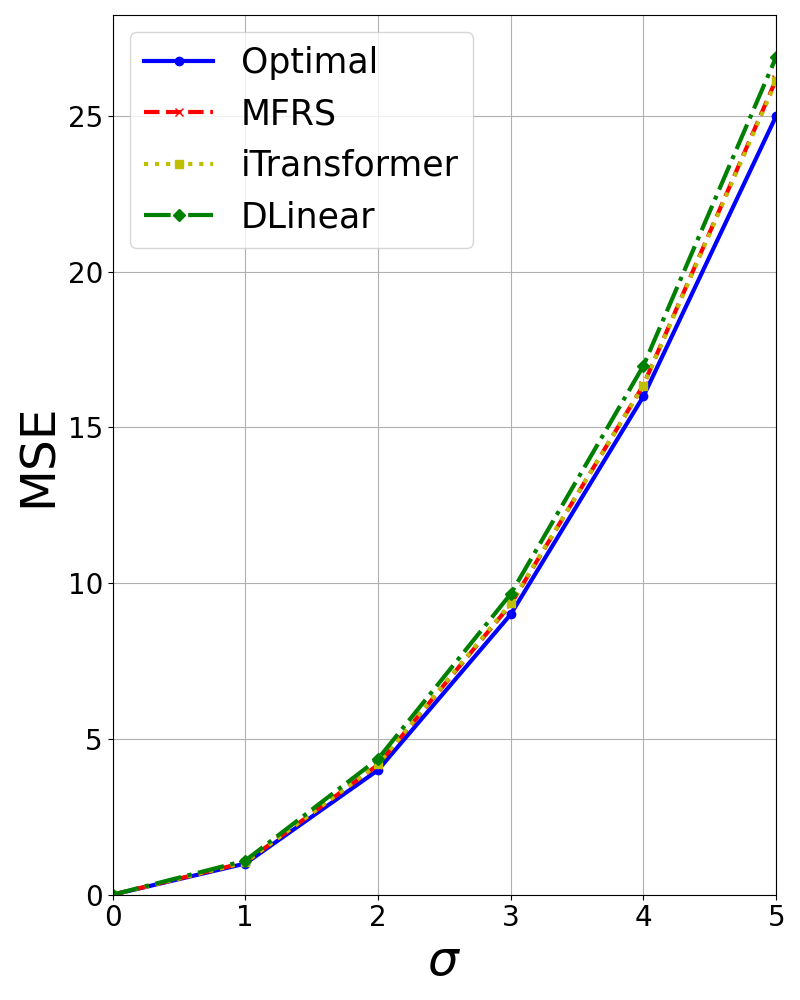}
    \end{minipage}
}
\subfigure[T~=~720]
{
    \begin{minipage}[b]{0.47\linewidth}
        \centering
        \includegraphics[width=\columnwidth]{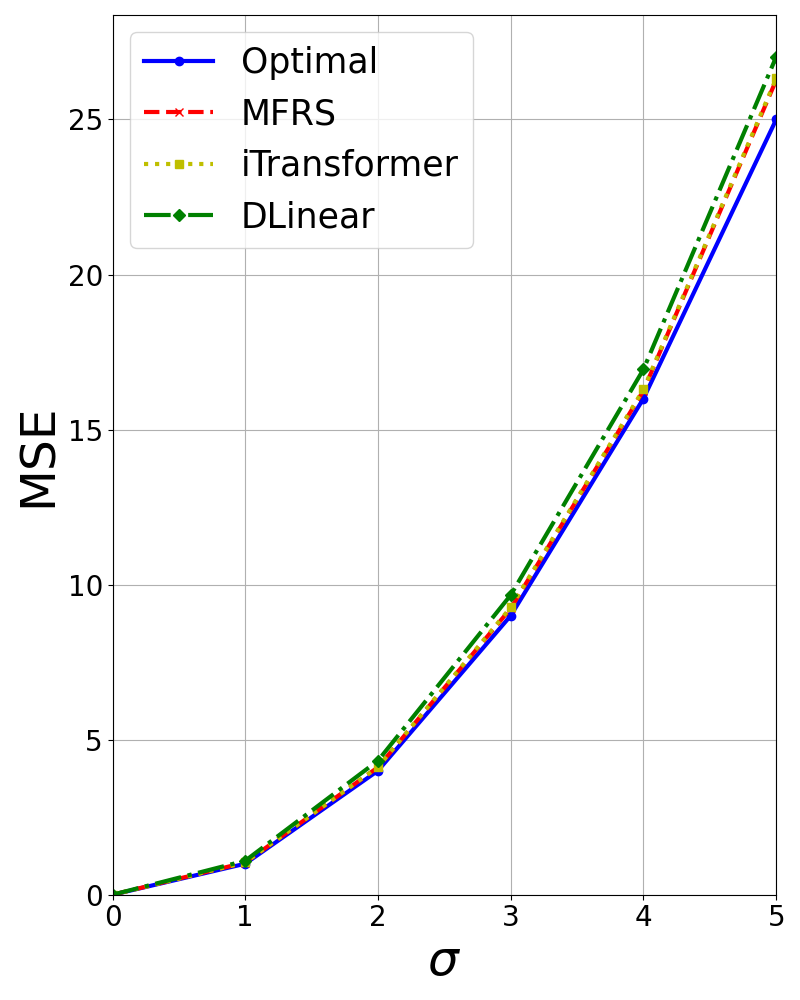}
    \end{minipage}
}
\vskip -0.1in
\caption{\textbf{Compose} with base-patterns $\mathcal{T} \in \left \{ 72, 36, 24, 18 \right \}$}
\label{figure-compose1}
\vskip -0.1in
\end{figure}

Figure~\ref{figure-compose1} presents performance comparison on \textbf{Compose} where periodic pattern $\mathcal{T}\in \left \{ 72, 36, 24, 18 \right \}$ is less than look-back window $S$. In these cases, at least one complete cycle can be observed in historical observations. It can be seen that MSE of the three models are already very close to the theoretical optimal values. Specifically, the performance of MFRS and iTransformer are almost identical and slightly superior to that of DLinea. Furthermore, they exhibit excellent performance in both short-term ($T=96$) and long-term prediction ($T=720$), without any deterioration almost as the prediction horizon increases.

\begin{figure}[h]
\vskip -0.1in
\centering
\subfigure[T~=~96]
{
    \begin{minipage}[b]{0.47\linewidth}
        \centering
        \includegraphics[width=\columnwidth]{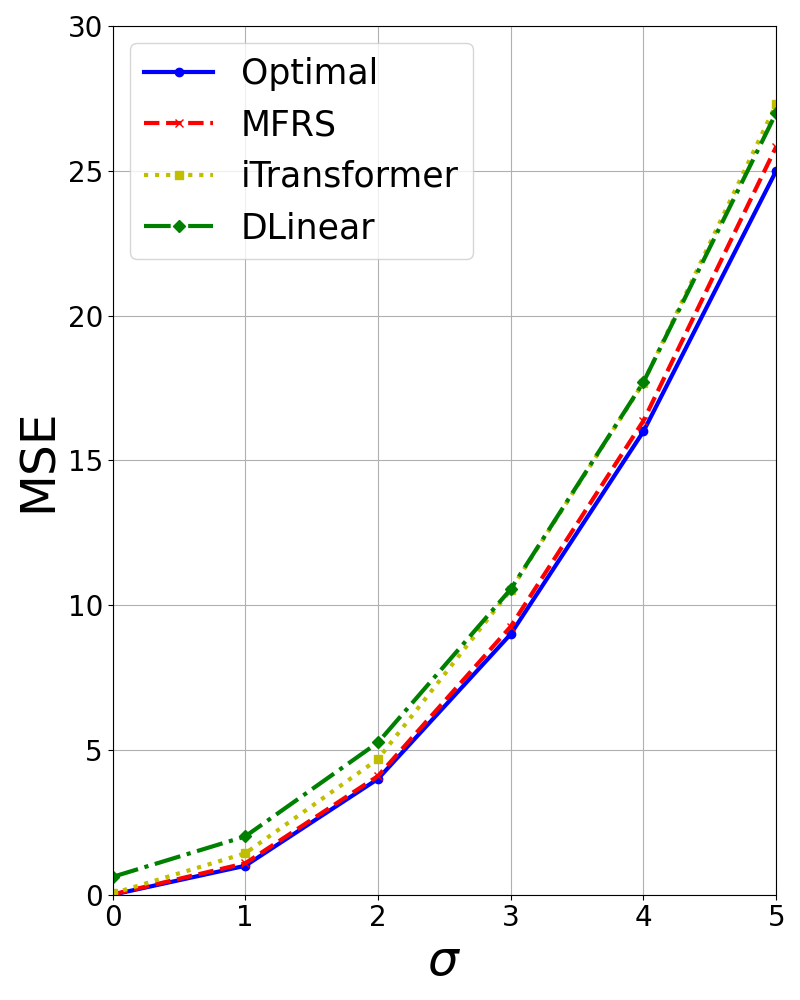}
    \end{minipage}
}
\subfigure[T~=~720]
{
    \begin{minipage}[b]{0.47\linewidth}
        \centering
        \includegraphics[width=\columnwidth]{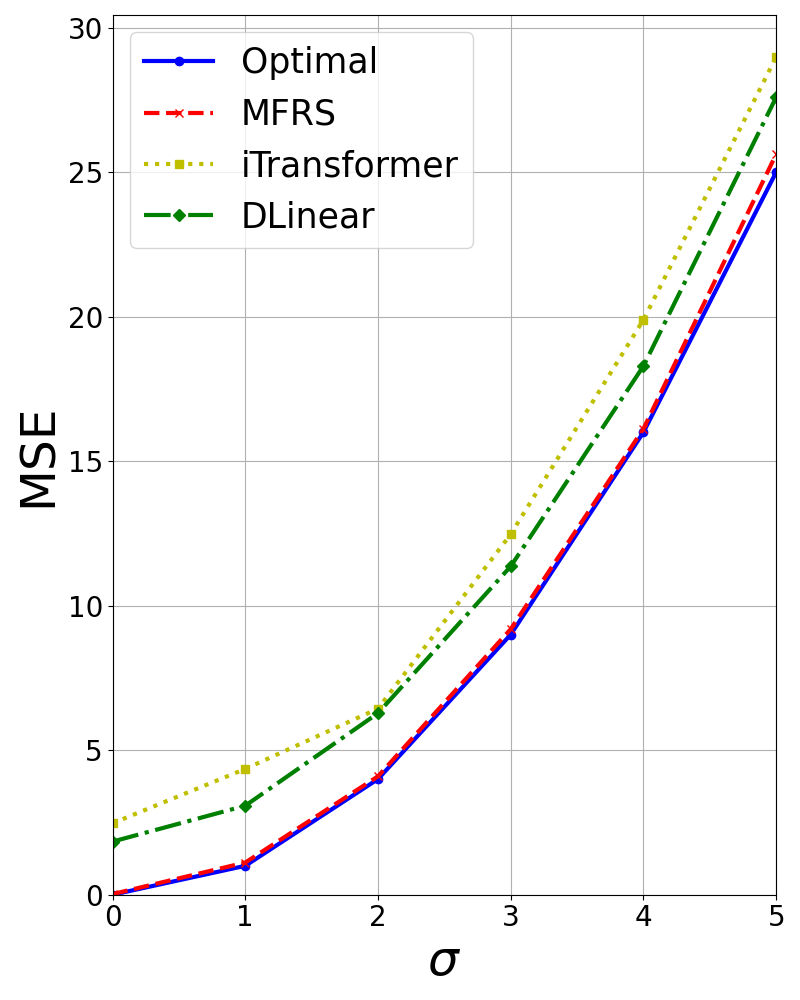}
    \end{minipage}
}
\vskip -0.1in
\caption{\textbf{Compose} with base-patterns $\mathcal{T} \in \left \{ 720, 360, 240, 180 \right \}$}
\label{figure-compose2}
\vskip -0.1in
\end{figure}

Figure~\ref{figure-compose2} shows experiment results on \textbf{Compose} where $\mathcal{T}\in \left \{ 720, 360, 240, 180 \right \}$ is larger than $S$. In this scenario, the performance of iTransformer and DLinear degrade significantly, especially in long-term prediction ($T=720$). However, MFRS remains unaffected, demonstrating stable performance in both short-term and long-term predictions and approaching the optimal value closely.

Based on the above analysis, it can be concluded that some current forcasters have already achieved theoretical optimality with sufficient historical observation. However, this scenario places higher demands on data and may not always be met, rendering the models less practical. In contrast, MFRS achieves optimality even with limited observations. The reason is that MFRS extracts periodic information contained within RS applying a further receptive field not restricted to look-back observations.

%%%%%%%%%%%%%%%%%%%%%%%%%%%%%%%%%%%%%%%%%%%%%%%%%%%%%%%%%%%%%%%%%%%%%%%%%%%%%%%
%%%%%%%%%%%%%%%%%%%%%%%%%%%%%%%%%%%%%%%%%%%%%%%%%%%%%%%%%%%%%%%%%%%%%%%%%%%%%%%
%                             Section 4.3
%%%%%%%%%%%%%%%%%%%%%%%%%%%%%%%%%%%%%%%%%%%%%%%%%%%%%%%%%%%%%%%%%%%%%%%%%%%%%%%
%%%%%%%%%%%%%%%%%%%%%%%%%%%%%%%%%%%%%%%%%%%%%%%%%%%%%%%%%%%%%%%%%%%%%%%%%%%%%%%
\subsection{Model Analysis}
\label{sec1-model}

\subsubsection{Impact of RS Types}
\label{sec2-types}

Apart from the common sine, waveforms such as sawtooth, rectangle, and pulse can all provide the same frequency information, and there is essentially no difference for them in serving as RS. To verify this statement, we implemented MFRS taking sawtooth, rectangle, and pulse as RS respectively, then conducted experiments on \textbf{Electricity} and \textbf{Weather}. All experimental settings were the same as in Section~\ref{sec1-open}. Finally, we took the average of the results for different prediction lengths. The metrics in Table~\ref{table-type-result} show that different types of RS hardly have any impact on the MFRS model.

\begin{table}[h]
\vskip -0.2in
\caption{results of different RS types}
\vskip 0.1in
\label{table-type-result}
\centering
\begin{tabular}{lcccc}
\toprule
\rowcolor[HTML]{FFFFFF} 
{\color[HTML]{333333} Datasets} & \multicolumn{2}{c}{\cellcolor[HTML]{FFFFFF}{\color[HTML]{333333} \textbf{Electricity}}} & \multicolumn{2}{c}{\cellcolor[HTML]{FFFFFF}{\color[HTML]{333333} \textbf{Weather}}} \\
\rowcolor[HTML]{F8F8F8} 
{\color[HTML]{333333} Metrics} & {\color[HTML]{333333} MSE} & {\color[HTML]{333333} MAE} & {\color[HTML]{333333} MSE} & {\color[HTML]{333333} MAE} \\
\midrule
\rowcolor[HTML]{FFFFFF} 
{\color[HTML]{333333} \textbf{sin}} & {\color[HTML]{333333} 0.162} & {\color[HTML]{333333} 0.256} & {\color[HTML]{333333} 0.244} & {\color[HTML]{333333} 0.270} \\
\rowcolor[HTML]{F8F8F8} 
{\color[HTML]{333333} \textbf{swatooth}} & {\color[HTML]{333333} 0.166} & {\color[HTML]{333333} 0.258} & {\color[HTML]{333333} 0.245} & {\color[HTML]{333333} 0.270} \\
\rowcolor[HTML]{FFFFFF} 
{\color[HTML]{333333} \textbf{reactangle}} & {\color[HTML]{333333} 0.166} & {\color[HTML]{333333} 0.256} & {\color[HTML]{333333} 0.245} & {\color[HTML]{333333} 0.270} \\
\rowcolor[HTML]{F8F8F8} 
{\color[HTML]{333333} \textbf{pulse}} & {\color[HTML]{333333} 0.165} & {\color[HTML]{333333} 0.257} & {\color[HTML]{333333} 0.246} & {\color[HTML]{333333} 0.271}\\
\bottomrule
\end{tabular}
\end{table}

\subsubsection{Impact of RS length}
\label{sec2-length}

$\mathbf{RS}_{t:t+S}$ primarily provides phase and frequency information in MFRS. The phase remains unchanged as time step $t$ is fixed, and the frequency has no concern with the length of RS. Given this, $\mathbf{RS}_{t:t+P}$ with the changed look-back window $P$ contains the same information, thus should have no impact on the performance of MFRS. To verify this statement experimentally, RS no longer share embedding layer with variates, but instead apply Embedding: $R^{P} \longrightarrow R^{D}$. All experimental settings remain the same as in Section~\ref{sec1-open}, except for the parameter $P$. Finally, we take the average of the results for different prediction lengths. As shown in Table~\ref{table-length-result}, there is no fundamental change in metrics whether the look-back window of RS is increased or decreased, which is consistent with our prior claim.

\begin{table}[h]
\caption{results of different RS length}
\vskip 0.1in
\label{table-length-result}
\centering
\begin{tabular}{cccccc}
\toprule
\multicolumn{3}{c}{\textbf{Electricity}} & \multicolumn{3}{c}{\textbf{Weather}} \\
P & MSE & MAE & P & MSE & MAE \\
\midrule
24 & 0.165 & 0.258 & 36 & 0.244 & 0.270 \\
48 & 0.165 & 0.258 & 72 & 0.245 & 0.270 \\
96 & 0.162 & 0.256 & 96 & 0.244 & 0.270 \\
168 & 0.165 & 0.258 & 144 & 0.244 & 0.270 \\
\bottomrule
\end{tabular}
\end{table}

\subsubsection{Shorter Look-back Length}
\label{sec2-window}

\begin{figure}[h]
\centering
\subfigure[\textbf{ETTm1}]
{
    \begin{minipage}[b]{\linewidth}
        \centering
        \includegraphics[width=\columnwidth]{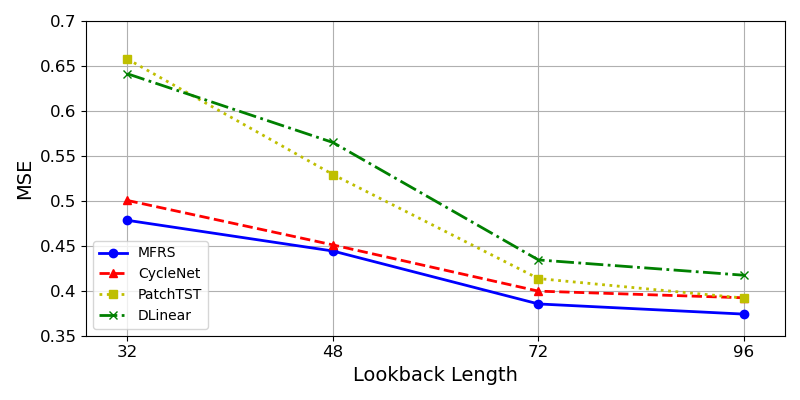}
    \end{minipage}
}
\subfigure[\textbf{ETTh1}]
{
    \begin{minipage}[b]{\linewidth}
        \centering
        \includegraphics[width=\columnwidth]{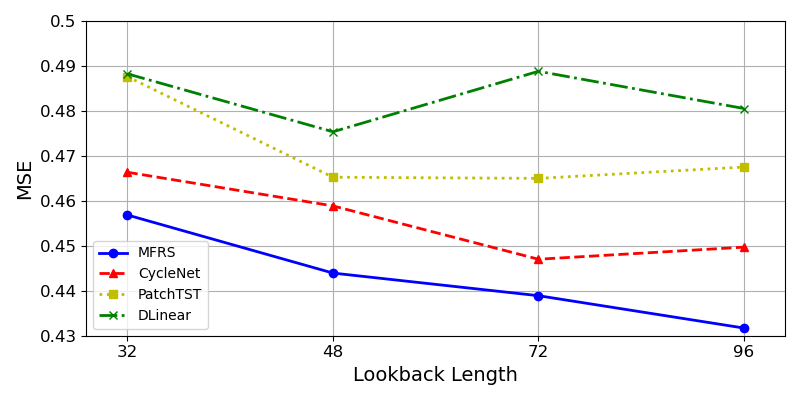}
    \end{minipage}
}
\vskip -0.1in
\caption{Performance on \textbf{ETTm1} and \textbf{ETTh1} with shorter look-back length $S\in \left \{ 96, 72, 48, 36 \right \}$ and prediction horizons $T\in \left \{ 96, 192, 336, 720 \right \}$. Results are averaged from all prediction horizons}
\label{figure-seq}
\end{figure}

A shorter look-back window contains less historical information, making it more difficult to capture long-term dependency, and thus degrading model performance. The efficient periodic pattern capturing capability of MFRS mitigates the adverse impact. Figure~\ref{figure-seq} illustrates the performance of MFRS, along with other outstanding models, such as CycleNet, PatchTST and DLinear, on \textbf{ETTm1} and \textbf{ETTh1} under less historical observation. \textbf{ETTm1} includes PBP $\mathcal{T} =96$, while \textbf{ETTh1} is derived from \textbf{ETTm1} through quarter downsampling, resulting in shorter PBP $\mathcal{T} =24$. 

The experimental results of the four models on dataset \textbf{ETTm1} clearly demonstrate two distinct patterns. MFRS and CycleNet exhibit a relatively gradual decline in performance as the look-back length decreases, while PatchTST and DLinear much steeper. This is because both MFRS and CycleNet incorporate periodic pattern into the modeling process. The difference lies in that the parameter $\mathcal{T} =96$ is manually set in CycleNet, whereas extracted automatically through spectral analysis in MFRS, thus possessing greater practicality. However, all models experience a similar slight decline on \textbf{ETTh1}. The reason is that \textbf{ETTm1} exhibits short-term period pattern, so fewer historical observations can provide relatively complete periodic information, which does not highlight the superiority of MFRS and CycleNet.

\section{Conclusion}
This paper visually confirms the existence of periodicity comprising several base-patterns within a time series and models this periodicity by capturing the correlations between the original series and reference series. The method extracts some frequency components resulting from periodicity by FFT, and generates the corresponding RS. Subsequently, an inverted Transformer structure model is employed to establish the dependency between variables and RS. As a novel approach, MFRS achieves state-of-the-art performance on real-world datasets across different domains. Furthermore, we have developed a set of synthetic datasets as new benchmarks, and experiments conducted on them demonstrate the efficiency of MFRS in extracting periodic features. Finally, we will focus on predicting trend changes to further enhance the model performance.

\nocite{langley00}

\bibliography{MFRS}
\bibliographystyle{icml2021}


\appendix
\begin{table*}[h]
\setlength{\tabcolsep}{6pt}
\renewcommand{\arraystretch}{1.5}
\label{table-datasets}
\centering
\caption{Datasets Information}
\vskip 0.15in
\begin{tabular}{|c|c|c|c|c|c|c|}
\hline
Dataset & \textbf{ETTh} & \textbf{ETTm} & \textbf{Electricity} & \textbf{Traffic} & \textbf{Weather} & \textbf{Solar} \\ \hline
Features & 7 & 7 & 321 & 862 & 21 & 137 \\ \hline
Timesteps & 17420 & 69680 & 26304 & 17544 & 52696 & 52560 \\ \hline
Frequency & Hourly & 15min & Hourly & Hourly & 10min & 10min \\ \hline
\end{tabular}
\end{table*}

\section{Channel Independence}
\label{app-channel}

There are two approaches to solving multi-channel time series forcasting tasks. One assumes that there exists some dependency among channels, which can be leveraged to enhance model performance, e.g., MTS-Mixers \cite{MTS-Mixers}, Crossformer \cite{Crossformer}. The other one completely disregards the dependency among channels, as other channels may provide redundant or even erroneous information, or such dependency is difficult to effectively establish, e.g., PatchTST \cite{PatchTST}. Our proposed model MFRS adopts the latter approach, namely Channel Independence, for which we provide a rigorous definition.

Let $\mathbf{Dataset}=\left [ \mathbf{X}^{\left ( 1 \right ) },..., \mathbf{X}^{\left ( C \right ) }  \right ]$ be the time series dataset with $C$ channels. Then split it into $C$ single-channel datasets $\mathbf{Dataset}_{i}=\left [ \mathbf{X}^{\left ( i \right ) }  \right ],i=1,...,C$. Suppose model $\mathbf{F}$ can handle input data with varying numbers of channels, meaning it is applicable to both MTSF and STSF. Let $\theta$ denote the model parameters of $\mathbf{F}$ after trained on $\mathbf{Dataset}$. The predicted values can then be expressed as

\begin{equation*}
    \begin{split}
        \left [ \mathbf{P}^{\left ( 1 \right ) },...,\mathbf{P}^{\left ( C \right ) }  \right ]=\mathbf{F}\left ( \left [ \mathbf{X}^{\left ( 1 \right ) },...,\mathbf{X}^{\left ( C \right ) } \right ] \mid \theta   \right )
    \end{split}
\end{equation*}

Where $\mathbf{P}^{\left ( i \right ) }$ represents the output of the $i$-th channel. On the other hand, the predicted values with $\mathbf{Dataset}_{i}$ as input can be expressed as

\begin{equation*}
    \begin{split}
        \left [ \mathbf{Q}^{\left ( i \right ) }\right ]=\mathbf{F}\left ( \left [ \mathbf{X}^{\left ( i \right ) } \right ] \mid \theta   \right )
    \end{split}
\end{equation*}

Model $\mathbf{F}$ can be considered channel independent if $\mathbf{P}^{\left ( i \right ) }=\mathbf{Q}^{\left ( i \right ) }$, equivalent to

\begin{equation*}
    \begin{split}
        &\mathbf{F}\left ( \left [ \mathbf{X}^{\left ( 1 \right ) },...,\mathbf{X}^{\left ( C \right ) } \right ] \mid \theta   \right ) 
    \end{split}    
\end{equation*}
\begin{equation*}
    \begin{split}
        = \left [\mathbf{F}\left ( \left [ \mathbf{X}^{\left ( 1 \right ) } \right ] \mid \theta   \right ),...,\mathbf{F}\left ( \left [ \mathbf{X}^{\left ( C \right ) } \right ] \mid \theta   \right ) \right ]
    \end{split}
\end{equation*}

In other words, $\mathbf{F}$ satisfies channel additivity.

\section{Method of Generating RS }
\label{app-refs}

Assume that the Primary Base-Patter $f$ extracted by module \textbf{BPE} contains $K-1$ harmonic components $2f,...,Kf$. The periodic pattern $\mathcal{T}=\frac{1}{f}$ is an integer, but its harmonic components may not be  necessarily. This point is crucial when generating RS, as it may involve modulo and exact division operations. Here, we express the generation method for single-cycle series of different waveforms (sine, sawtooth, rectangle, and pulse) using the following four formulas, all of which can serve as RS containing periodic information.

\begin{equation*}
    \begin{split}
        \mathbf{Sine}_{t}^{\left ( k \right ) } &=sin\left ( 2\pi \ast kf\ast t \right )\\
        \mathbf{Sawtooth}_{t}^{\left ( k \right ) } &=\left ( t\ast k \right ) \%  \left ( \mathcal{T}  \right )\\
        \mathbf{Rectangle}_{t}^{\left ( k \right ) } &=\left \lfloor \frac{ 2\ast k\ast t}{\mathcal{T} }  \right \rfloor \% 2\\
        \mathbf{Pulse}_{t}^{\left ( k \right ) } &=
        \begin{cases}
            1\text{,~~~~if~~} \left ( t\ast k \right ) \% \mathcal{T} = 0\\
            0\text{,~~~~else}
        \end{cases}
    \end{split}
\end{equation*}

Where $k=1,...,K$ and $\%$ denotes modulo operation.

\section{Open Datasets}
\label{app-open-datasets}

\subsection{Description}
\label{app1-description}

The real-world datasets mentioned in Section~\ref{sec1-datasets} can be roughly divided into two categories: social field involving human activities (e.g, \textbf{Traffic}) and natural field without human intervention (e.g, \textbf{Weather}). The relevant parameters of them are exhibited in Table~\ref{table-datasets}, where \textit{Features} represents the number of channels, \textit{Timesteps} indicates the number of sampling points, and \textit{Frequency} denotes the interval between time steps.

\subsection{Abundant spectrum}
\label{app1-spectrum}

A real-world dataset typically contains similar periodic patterns across channels, which is the reason why all channels share a set of RS in MFRS. Furthermore, the patterns are determined by the sampling interval between time points to a large extent. We performed FFT operations on three channels from each of (\textbf{Weather}, \textbf{ETTm1}, \textbf{Traffic}) with different intervals: 10 minutes, 15 minutes and 1 hour. All the spectrum obtained are then converted by the method presented in Section~\ref{sec2-bpe} to extract PBP.

\begin{figure}[h]
\begin{center}
\centerline
{
    \begin{minipage}[b]{\linewidth}
        \centering
        \includegraphics[width=\columnwidth]{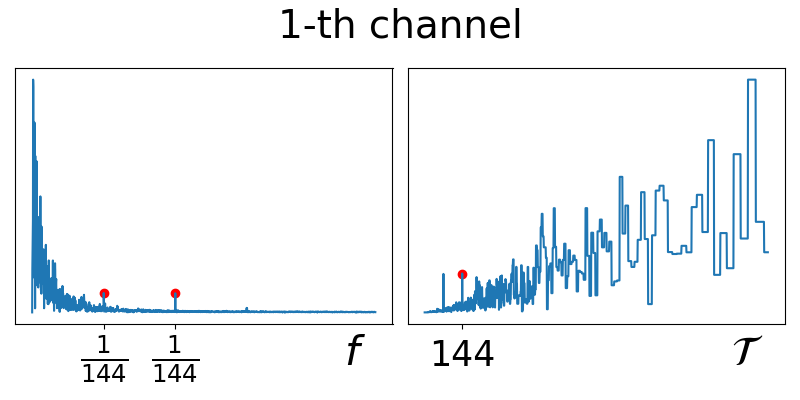}
    \end{minipage}
}
{
    \begin{minipage}[b]{\linewidth}
        \centering
        \includegraphics[width=\columnwidth]{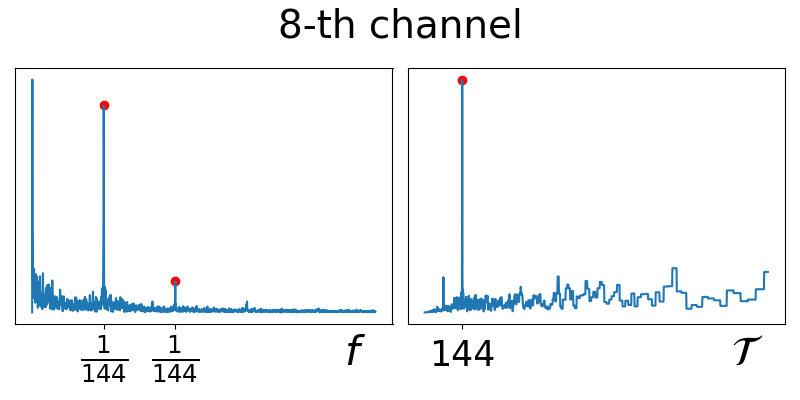}
    \end{minipage}
}
{
    \begin{minipage}[b]{\linewidth}
        \centering
        \includegraphics[width=\columnwidth]{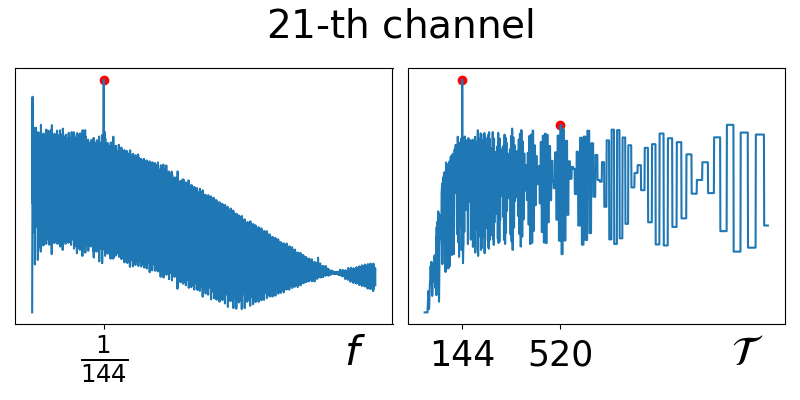}
    \end{minipage}
}
\caption{spectrum of \textbf{Weather}}    
\label{figure-weather}
\end{center}
\vskip -0.2in
\end{figure}

As shown in Figure~\ref{figure-weather}, the PBP $\mathcal{T}=144$ corresponds exactly to one day. However, \textbf{Weather} does not have a weekly periodic pattern because the natural world, without human intervention, does not distinguish between weekdays and weekends. And we get an abnormal PBP $\mathcal{T}=144$ from the converted spectrum of $21$-th channel, which is caused by a large amount of noise.

\begin{figure}[h]
\begin{center}
\centerline
{
    \begin{minipage}[b]{\linewidth}
        \centering
        \includegraphics[width=\columnwidth]{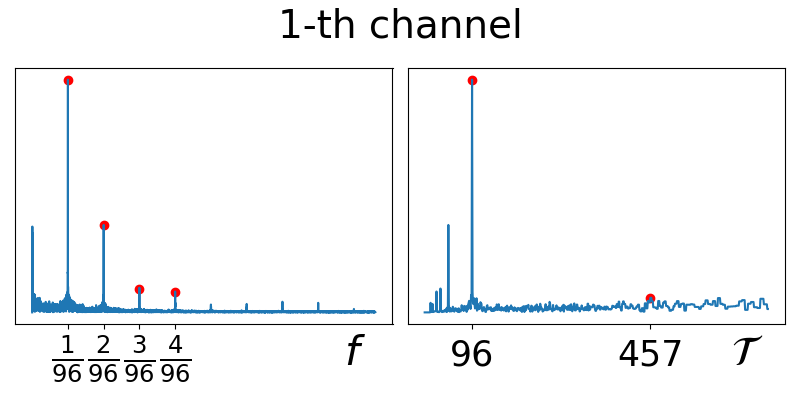}
    \end{minipage}
}
{
    \begin{minipage}[b]{\linewidth}
        \centering
        \includegraphics[width=\columnwidth]{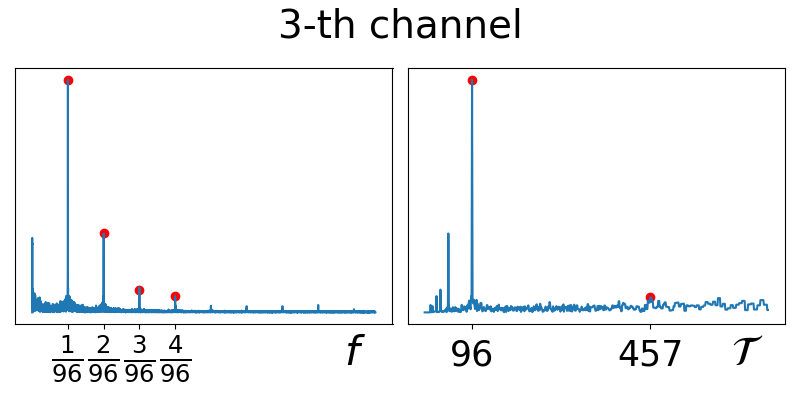}
    \end{minipage}
}
{
    \begin{minipage}[b]{\linewidth}
        \centering
        \includegraphics[width=\columnwidth]{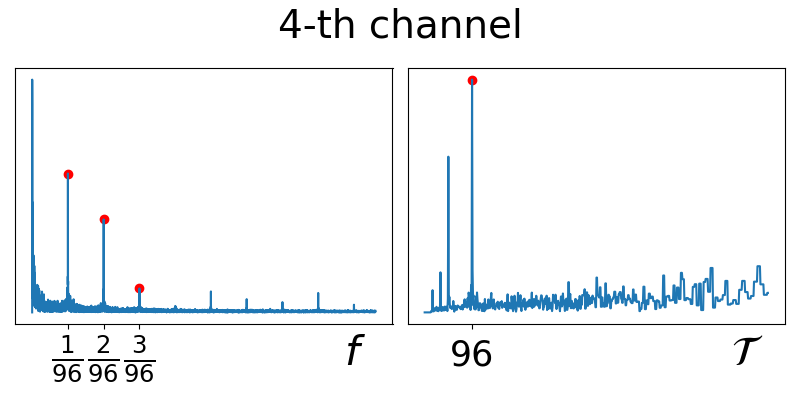}
    \end{minipage}
}
\caption{spectrum of \textbf{ETTm1}}    
\label{figure-ettm1}
\end{center}
\vskip -0.2in
\end{figure}

As shown in Figure~\ref{figure-ettm1}, the PBP $\mathcal{T}=96$ corresponds exactly to one day. Additionally, an value $\mathcal{T}=457$ is obtained from the converted spectrum of both $1$-th and $21$-th channel, but not found in $4$-th channel.

\begin{figure}[h]
\begin{center}
\centerline
{
    \begin{minipage}[b]{\linewidth}
        \centering
        \includegraphics[width=\columnwidth]{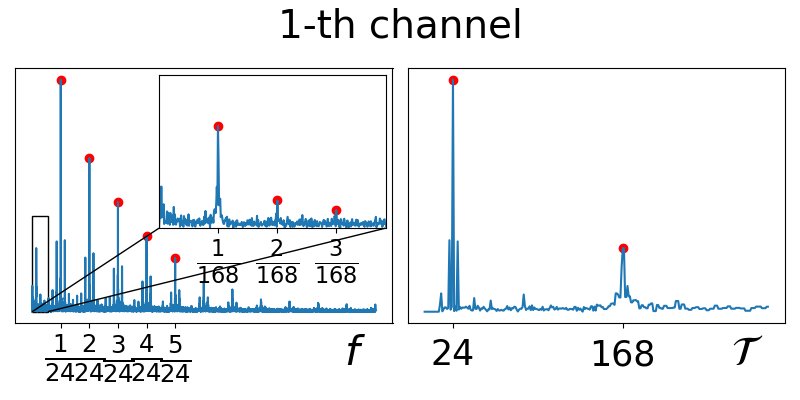}
    \end{minipage}
}
{
    \begin{minipage}[b]{\linewidth}
        \centering
        \includegraphics[width=\columnwidth]{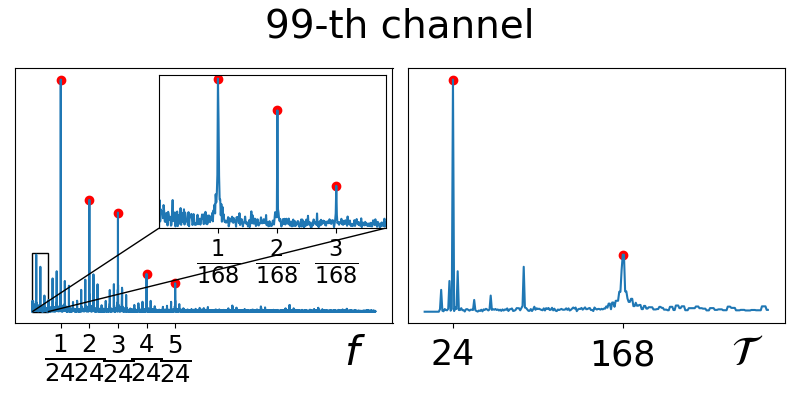}
    \end{minipage}
}
{
    \begin{minipage}[b]{\linewidth}
        \centering
        \includegraphics[width=\columnwidth]{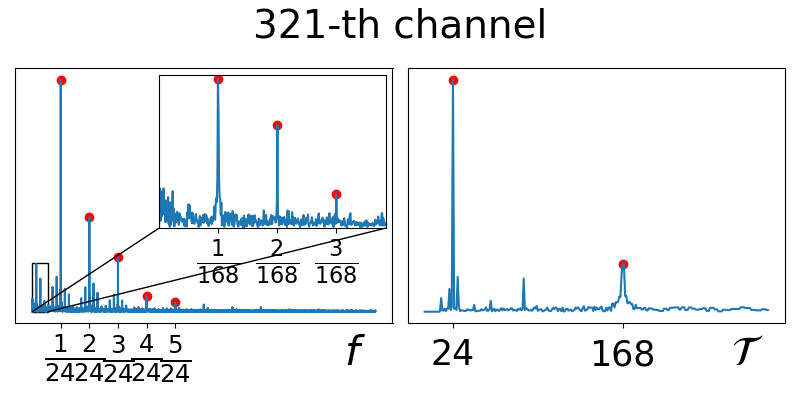}
    \end{minipage}
}
\caption{spectrum of \textbf{Traffic}}    
\label{figure-traffic}
\end{center}
\vskip -0.2in
\end{figure}

As shown in Figure~\ref{figure-traffic}, PBP of \textbf{Traffic} include $\mathcal{T}_{1}=24$ and $\mathcal{T}_{2}=168=24\times 7$, indicating hourly and weekly periodic patterns. This aligns with empirical knowledge, that traffic flow tends to exhibit similar variation each day, such as morning and evening rush hours, and differ between weekdays and weekends.

\subsection{Full Results}
\label{app2-full-results}

% \usepackage{multirow}
% \usepackage[normalem]{ulem}
\begin{table*}[htbp]\small
\setlength{\tabcolsep}{3pt}
\renewcommand{\arraystretch}{1.8}

\caption{Detailed results on open datasets with fixed look-back length $S=96$.}
\label{table_results_all}
\vskip 0.2in
\centering
\begin{tabular}{cc|cccccccccccccccc}
\hline
\multicolumn{2}{c|}{models} & \multicolumn{2}{c}{MFRS} & \multicolumn{2}{c}{CycleNet} & \multicolumn{2}{c}{iTransformer} & \multicolumn{2}{c}{PatchTST} & \multicolumn{2}{c}{DLinear} & \multicolumn{2}{c}{FEDformer} & \multicolumn{2}{c}{TimesNet} & \multicolumn{2}{c}{TiDE} \\
\multicolumn{2}{c|}{metrics} & \textbf{MSE} & \textbf{MAE} & \textbf{MSE} & \textbf{MAE} & \textbf{MSE} & \textbf{MAE} & \textbf{MSE} & \textbf{MAE} & \textbf{MSE} & \textbf{MAE} & \textbf{MSE} & \textbf{MAE} & \textbf{MSE} & \textbf{MAE} & \textbf{MSE} & \textbf{MAE} \\ \hline
 & 96 & \cellcolor[HTML]{EFEFEF}\textbf{0.306} & {\ul 0.348} & \cellcolor[HTML]{EFEFEF}{\ul 0.319} & \textbf{0.36} & \cellcolor[HTML]{EFEFEF}0.334 & 0.368 & \cellcolor[HTML]{EFEFEF}0.329 & 0.367 & \cellcolor[HTML]{EFEFEF}0.345 & 0.372 & \cellcolor[HTML]{EFEFEF}0.379 & 0.419 & \cellcolor[HTML]{EFEFEF}0.338 & 0.375 & \cellcolor[HTML]{EFEFEF}0.364 & 0.387 \\
 & 192 & \cellcolor[HTML]{EFEFEF}\textbf{0.354} & \textbf{0.377} & \cellcolor[HTML]{EFEFEF}{\ul 0.36} & {\ul 0.381} & \cellcolor[HTML]{EFEFEF}0.377 & 0.391 & \cellcolor[HTML]{EFEFEF}0.367 & 0.385 & \cellcolor[HTML]{EFEFEF}0.38 & 0.389 & \cellcolor[HTML]{EFEFEF}0.426 & 0.441 & \cellcolor[HTML]{EFEFEF}0.374 & 0.387 & \cellcolor[HTML]{EFEFEF}0.398 & 0.404 \\
 & 336 & \cellcolor[HTML]{EFEFEF}\textbf{0.386} & \textbf{0.4} & \cellcolor[HTML]{EFEFEF}{\ul 0.389} & {\ul 0.403} & \cellcolor[HTML]{EFEFEF}0.426 & 0.42 & \cellcolor[HTML]{EFEFEF}0.399 & 0.41 & \cellcolor[HTML]{EFEFEF}0.413 & 0.413 & \cellcolor[HTML]{EFEFEF}0.445 & 0.459 & \cellcolor[HTML]{EFEFEF}0.41 & 0.411 & \cellcolor[HTML]{EFEFEF}0.428 & 0.425 \\
\multirow{-4}{*}{\textbf{ETTm1}} & 720 & \cellcolor[HTML]{EFEFEF}\textbf{0.443} & \textbf{0.436} & \cellcolor[HTML]{EFEFEF}{\ul 0.447} & 0.441 & \cellcolor[HTML]{EFEFEF}0.491 & 0.459 & \cellcolor[HTML]{EFEFEF}0.454 & {\ul 0.439} & \cellcolor[HTML]{EFEFEF}0.474 & 0.453 & \cellcolor[HTML]{EFEFEF}0.543 & 0.49 & \cellcolor[HTML]{EFEFEF}0.478 & 0.45 & \cellcolor[HTML]{EFEFEF}0.487 & 0.461 \\ \hline
 & 96 & \cellcolor[HTML]{EFEFEF}0.177 & {\ul 0.257} & \cellcolor[HTML]{EFEFEF}\textbf{0.163} & \textbf{0.246} & \cellcolor[HTML]{EFEFEF}0.18 & 0.264 & \cellcolor[HTML]{EFEFEF}{\ul 0.175} & 0.259 & \cellcolor[HTML]{EFEFEF}0.193 & 0.292 & \cellcolor[HTML]{EFEFEF}0.203 & 0.287 & \cellcolor[HTML]{EFEFEF}0.187 & 0.267 & \cellcolor[HTML]{EFEFEF}0.207 & 0.305 \\
 & 192 & \cellcolor[HTML]{EFEFEF}0.244 & {\ul 0.302} & \cellcolor[HTML]{EFEFEF}\textbf{0.229} & \textbf{0.29} & \cellcolor[HTML]{EFEFEF}0.25 & 0.309 & \cellcolor[HTML]{EFEFEF}{\ul 0.241} & 0.302 & \cellcolor[HTML]{EFEFEF}0.284 & 0.362 & \cellcolor[HTML]{EFEFEF}0.269 & 0.328 & \cellcolor[HTML]{EFEFEF}0.249 & 0.309 & \cellcolor[HTML]{EFEFEF}0.29 & 0.364 \\
 & 336 & \cellcolor[HTML]{EFEFEF}{\ul 0.298} & {\ul 0.339} & \cellcolor[HTML]{EFEFEF}\textbf{0.284} & \textbf{0.327} & \cellcolor[HTML]{EFEFEF}0.311 & 0.348 & \cellcolor[HTML]{EFEFEF}0.305 & 0.343 & \cellcolor[HTML]{EFEFEF}0.369 & 0.427 & \cellcolor[HTML]{EFEFEF}0.325 & 0.366 & \cellcolor[HTML]{EFEFEF}0.321 & 0.351 & \cellcolor[HTML]{EFEFEF}0.377 & 0.422 \\
\multirow{-4}{*}{\textbf{ETTm2}} & 720 & \cellcolor[HTML]{EFEFEF}\textbf{0.384} & {\ul 0.395} & \cellcolor[HTML]{EFEFEF}{\ul 0.389} & \textbf{0.391} & \cellcolor[HTML]{EFEFEF}0.412 & 0.407 & \cellcolor[HTML]{EFEFEF}0.402 & 0.4 & \cellcolor[HTML]{EFEFEF}0.554 & 0.522 & \cellcolor[HTML]{EFEFEF}0.421 & 0.415 & \cellcolor[HTML]{EFEFEF}0.408 & 0.403 & \cellcolor[HTML]{EFEFEF}0.558 & 0.524 \\ \hline
 & 96 & \cellcolor[HTML]{EFEFEF}\textbf{0.371} & \textbf{0.395} & \cellcolor[HTML]{EFEFEF}{\ul 0.375} & {\ul 0.395} & \cellcolor[HTML]{EFEFEF}0.386 & 0.405 & \cellcolor[HTML]{EFEFEF}0.414 & 0.419 & \cellcolor[HTML]{EFEFEF}0.386 & 0.4 & \cellcolor[HTML]{EFEFEF}0.376 & 0.419 & \cellcolor[HTML]{EFEFEF}0.384 & 0.402 & \cellcolor[HTML]{EFEFEF}0.479 & 0.464 \\
 & 192 & \cellcolor[HTML]{EFEFEF}{\ul 0.421} & \textbf{0.424} & \cellcolor[HTML]{EFEFEF}0.436 & {\ul 0.428} & \cellcolor[HTML]{EFEFEF}0.441 & 0.436 & \cellcolor[HTML]{EFEFEF}0.46 & 0.445 & \cellcolor[HTML]{EFEFEF}0.437 & 0.432 & \cellcolor[HTML]{EFEFEF}\textbf{0.42} & 0.448 & \cellcolor[HTML]{EFEFEF}0.436 & 0.429 & \cellcolor[HTML]{EFEFEF}0.525 & 0.492 \\
 & 336 & \cellcolor[HTML]{EFEFEF}{\ul 0.461} & \textbf{0.442} & \cellcolor[HTML]{EFEFEF}0.496 & {\ul 0.455} & \cellcolor[HTML]{EFEFEF}0.487 & 0.458 & \cellcolor[HTML]{EFEFEF}0.501 & 0.466 & \cellcolor[HTML]{EFEFEF}0.481 & 0.459 & \cellcolor[HTML]{EFEFEF}\textbf{0.459} & 0.465 & \cellcolor[HTML]{EFEFEF}0.491 & 0.469 & \cellcolor[HTML]{EFEFEF}0.565 & 0.515 \\
\multirow{-4}{*}{\textbf{ETTh1}} & 720 & \cellcolor[HTML]{EFEFEF}\textbf{0.461} & \textbf{0.462} & \cellcolor[HTML]{EFEFEF}0.52 & {\ul 0.484} & \cellcolor[HTML]{EFEFEF}0.503 & 0.491 & \cellcolor[HTML]{EFEFEF}{\ul 0.5} & 0.488 & \cellcolor[HTML]{EFEFEF}0.519 & 0.516 & \cellcolor[HTML]{EFEFEF}0.506 & 0.507 & \cellcolor[HTML]{EFEFEF}0.521 & 0.5 & \cellcolor[HTML]{EFEFEF}0.594 & 0.558 \\ \hline
 & 96 & \cellcolor[HTML]{EFEFEF}\textbf{0.291} & \textbf{0.34} & \cellcolor[HTML]{EFEFEF}0.298 & {\ul 0.344} & \cellcolor[HTML]{EFEFEF}{\ul 0.297} & 0.349 & \cellcolor[HTML]{EFEFEF}0.302 & 0.348 & \cellcolor[HTML]{EFEFEF}0.333 & 0.387 & \cellcolor[HTML]{EFEFEF}0.358 & 0.397 & \cellcolor[HTML]{EFEFEF}0.34 & 0.374 & \cellcolor[HTML]{EFEFEF}0.4 & 0.44 \\
 & 192 & \cellcolor[HTML]{EFEFEF}\textbf{0.361} & \textbf{0.387} & \cellcolor[HTML]{EFEFEF}{\ul 0.372} & {\ul 0.396} & \cellcolor[HTML]{EFEFEF}0.38 & 0.4 & \cellcolor[HTML]{EFEFEF}0.388 & 0.4 & \cellcolor[HTML]{EFEFEF}0.477 & 0.476 & \cellcolor[HTML]{EFEFEF}0.429 & 0.439 & \cellcolor[HTML]{EFEFEF}0.402 & 0.414 & \cellcolor[HTML]{EFEFEF}0.528 & 0.509 \\
 & 336 & \cellcolor[HTML]{EFEFEF}\textbf{0.41} & \textbf{0.42} & \cellcolor[HTML]{EFEFEF}0.431 & 0.439 & \cellcolor[HTML]{EFEFEF}0.428 & {\ul 0.432} & \cellcolor[HTML]{EFEFEF}{\ul 0.426} & 0.433 & \cellcolor[HTML]{EFEFEF}0.594 & 0.541 & \cellcolor[HTML]{EFEFEF}0.496 & 0.487 & \cellcolor[HTML]{EFEFEF}0.452 & 0.452 & \cellcolor[HTML]{EFEFEF}0.643 & 0.571 \\
\multirow{-4}{*}{\textbf{ETTh2}} & 720 & \cellcolor[HTML]{EFEFEF}\textbf{0.414} & \textbf{0.439} & \cellcolor[HTML]{EFEFEF}0.45 & 0.458 & \cellcolor[HTML]{EFEFEF}{\ul 0.427} & {\ul 0.445} & \cellcolor[HTML]{EFEFEF}0.431 & 0.446 & \cellcolor[HTML]{EFEFEF}0.831 & 0.657 & \cellcolor[HTML]{EFEFEF}0.463 & 0.474 & \cellcolor[HTML]{EFEFEF}0.462 & 0.468 & \cellcolor[HTML]{EFEFEF}0.874 & 0.697 \\ \hline
 & 96 & \cellcolor[HTML]{EFEFEF}\textbf{0.132} & \textbf{0.227} & \cellcolor[HTML]{EFEFEF}{\ul 0.136} & {\ul 0.229} & \cellcolor[HTML]{EFEFEF}0.148 & 0.24 & \cellcolor[HTML]{EFEFEF}0.181 & 0.27 & \cellcolor[HTML]{EFEFEF}0.197 & 0.282 & \cellcolor[HTML]{EFEFEF}0.193 & 0.308 & \cellcolor[HTML]{EFEFEF}0.168 & 0.272 & \cellcolor[HTML]{EFEFEF}0.237 & 0.329 \\
 & 192 & \cellcolor[HTML]{EFEFEF}\textbf{0.15} & \textbf{0.244} & \cellcolor[HTML]{EFEFEF}{\ul 0.152} & {\ul 0.244} & \cellcolor[HTML]{EFEFEF}0.162 & 0.253 & \cellcolor[HTML]{EFEFEF}0.188 & 0.274 & \cellcolor[HTML]{EFEFEF}0.196 & 0.285 & \cellcolor[HTML]{EFEFEF}0.201 & 0.315 & \cellcolor[HTML]{EFEFEF}0.184 & 0.289 & \cellcolor[HTML]{EFEFEF}0.236 & 0.33 \\
 & 336 & \cellcolor[HTML]{EFEFEF}\textbf{0.167} & \textbf{0.263} & \cellcolor[HTML]{EFEFEF}{\ul 0.17} & {\ul 0.264} & \cellcolor[HTML]{EFEFEF}0.178 & 0.269 & \cellcolor[HTML]{EFEFEF}0.204 & 0.293 & \cellcolor[HTML]{EFEFEF}0.209 & 0.301 & \cellcolor[HTML]{EFEFEF}0.214 & 0.329 & \cellcolor[HTML]{EFEFEF}0.198 & 0.3 & \cellcolor[HTML]{EFEFEF}0.249 & 0.344 \\
\multirow{-4}{*}{\textbf{\begin{tabular}[c]{@{}c@{}}Eelec-\\ tricity\end{tabular}}} & 720 & \cellcolor[HTML]{EFEFEF}\textbf{0.195} & \textbf{0.29} & \cellcolor[HTML]{EFEFEF}{\ul 0.212} & {\ul 0.299} & \cellcolor[HTML]{EFEFEF}0.225 & 0.317 & \cellcolor[HTML]{EFEFEF}0.246 & 0.324 & \cellcolor[HTML]{EFEFEF}0.245 & 0.333 & \cellcolor[HTML]{EFEFEF}0.246 & 0.355 & \cellcolor[HTML]{EFEFEF}0.22 & 0.32 & \cellcolor[HTML]{EFEFEF}0.284 & 0.373 \\ \hline
 & 96 & \cellcolor[HTML]{EFEFEF}\textbf{0.384} & \textbf{0.259} & \cellcolor[HTML]{EFEFEF}0.458 & 0.296 & \cellcolor[HTML]{EFEFEF}{\ul 0.395} & {\ul 0.268} & \cellcolor[HTML]{EFEFEF}0.462 & 0.295 & \cellcolor[HTML]{EFEFEF}0.65 & 0.396 & \cellcolor[HTML]{EFEFEF}0.587 & 0.366 & \cellcolor[HTML]{EFEFEF}0.593 & 0.321 & \cellcolor[HTML]{EFEFEF}0.805 & 0.493 \\
 & 192 & \cellcolor[HTML]{EFEFEF}\textbf{0.398} & \textbf{0.264} & \cellcolor[HTML]{EFEFEF}0.457 & 0.294 & \cellcolor[HTML]{EFEFEF}{\ul 0.417} & {\ul 0.276} & \cellcolor[HTML]{EFEFEF}0.466 & 0.296 & \cellcolor[HTML]{EFEFEF}0.598 & 0.37 & \cellcolor[HTML]{EFEFEF}0.604 & 0.373 & \cellcolor[HTML]{EFEFEF}0.617 & 0.336 & \cellcolor[HTML]{EFEFEF}0.756 & 0.474 \\
 & 336 & \cellcolor[HTML]{EFEFEF}\textbf{0.414} & \textbf{0.272} & \cellcolor[HTML]{EFEFEF}0.47 & 0.299 & \cellcolor[HTML]{EFEFEF}{\ul 0.433} & {\ul 0.283} & \cellcolor[HTML]{EFEFEF}0.482 & 0.304 & \cellcolor[HTML]{EFEFEF}0.605 & 0.373 & \cellcolor[HTML]{EFEFEF}0.621 & 0.383 & \cellcolor[HTML]{EFEFEF}0.629 & 0.336 & \cellcolor[HTML]{EFEFEF}0.762 & 0.477 \\
\multirow{-4}{*}{\textbf{Traffic}} & 720 & \cellcolor[HTML]{EFEFEF}\textbf{0.44} & \textbf{0.286} & \cellcolor[HTML]{EFEFEF}0.502 & 0.314 & \cellcolor[HTML]{EFEFEF}{\ul 0.467} & {\ul 0.302} & \cellcolor[HTML]{EFEFEF}0.514 & 0.322 & \cellcolor[HTML]{EFEFEF}0.645 & 0.394 & \cellcolor[HTML]{EFEFEF}0.626 & 0.382 & \cellcolor[HTML]{EFEFEF}0.64 & 0.35 & \cellcolor[HTML]{EFEFEF}0.719 & 0.449 \\ \hline
 & 96 & \cellcolor[HTML]{EFEFEF}\textbf{0.156} & \textbf{0.199} & \cellcolor[HTML]{EFEFEF}{\ul 0.158} & {\ul 0.203} & \cellcolor[HTML]{EFEFEF}0.174 & 0.214 & \cellcolor[HTML]{EFEFEF}0.177 & 0.218 & \cellcolor[HTML]{EFEFEF}0.196 & 0.255 & \cellcolor[HTML]{EFEFEF}0.217 & 0.296 & \cellcolor[HTML]{EFEFEF}0.172 & 0.22 & \cellcolor[HTML]{EFEFEF}0.202 & 0.261 \\
 & 192 & \cellcolor[HTML]{EFEFEF}\textbf{0.207} & \textbf{0.247} & \cellcolor[HTML]{EFEFEF}{\ul 0.207} & {\ul 0.247} & \cellcolor[HTML]{EFEFEF}0.221 & 0.254 & \cellcolor[HTML]{EFEFEF}0.225 & 0.259 & \cellcolor[HTML]{EFEFEF}0.237 & 0.296 & \cellcolor[HTML]{EFEFEF}0.276 & 0.336 & \cellcolor[HTML]{EFEFEF}0.219 & 0.261 & \cellcolor[HTML]{EFEFEF}0.242 & 0.298 \\
 & 336 & \cellcolor[HTML]{EFEFEF}{\ul 0.263} & \textbf{0.288} & \cellcolor[HTML]{EFEFEF}\textbf{0.262} & {\ul 0.289} & \cellcolor[HTML]{EFEFEF}0.278 & 0.296 & \cellcolor[HTML]{EFEFEF}0.278 & 0.297 & \cellcolor[HTML]{EFEFEF}0.283 & 0.335 & \cellcolor[HTML]{EFEFEF}0.339 & 0.38 & \cellcolor[HTML]{EFEFEF}0.28 & 0.306 & \cellcolor[HTML]{EFEFEF}0.287 & 0.335 \\
\multirow{-4}{*}{\textbf{Weather}} & 720 & \cellcolor[HTML]{EFEFEF}\textbf{0.344} & \textbf{0.341} & \cellcolor[HTML]{EFEFEF}{\ul 0.344} & {\ul 0.344} & \cellcolor[HTML]{EFEFEF}0.358 & 0.347 & \cellcolor[HTML]{EFEFEF}0.354 & 0.348 & \cellcolor[HTML]{EFEFEF}0.345 & 0.381 & \cellcolor[HTML]{EFEFEF}0.403 & 0.428 & \cellcolor[HTML]{EFEFEF}0.365 & 0.359 & \cellcolor[HTML]{EFEFEF}0.351 & 0.386 \\ \hline
 & 96 & \cellcolor[HTML]{EFEFEF}{\ul 0.191} & \textbf{0.236} & \cellcolor[HTML]{EFEFEF}\textbf{0.19} & 0.247 & \cellcolor[HTML]{EFEFEF}0.203 & {\ul 0.237} & \cellcolor[HTML]{EFEFEF}0.234 & 0.286 & \cellcolor[HTML]{EFEFEF}0.29 & 0.378 & \cellcolor[HTML]{EFEFEF}0.242 & 0.342 & \cellcolor[HTML]{EFEFEF}0.25 & 0.292 & \cellcolor[HTML]{EFEFEF}0.312 & 0.399 \\
 & 192 & \cellcolor[HTML]{EFEFEF}{\ul 0.225} & \textbf{0.256} & \cellcolor[HTML]{EFEFEF}\textbf{0.21} & 0.266 & \cellcolor[HTML]{EFEFEF}0.233 & {\ul 0.261} & \cellcolor[HTML]{EFEFEF}0.267 & 0.31 & \cellcolor[HTML]{EFEFEF}0.32 & 0.398 & \cellcolor[HTML]{EFEFEF}0.285 & 0.38 & \cellcolor[HTML]{EFEFEF}0.296 & 0.318 & \cellcolor[HTML]{EFEFEF}0.339 & 0.416 \\
 & 336 & \cellcolor[HTML]{EFEFEF}{\ul 0.241} & {\ul 0.268} & \cellcolor[HTML]{EFEFEF}\textbf{0.217} & \textbf{0.266} & \cellcolor[HTML]{EFEFEF}0.248 & 0.273 & \cellcolor[HTML]{EFEFEF}0.29 & 0.315 & \cellcolor[HTML]{EFEFEF}0.353 & 0.415 & \cellcolor[HTML]{EFEFEF}0.282 & 0.376 & \cellcolor[HTML]{EFEFEF}0.319 & 0.33 & \cellcolor[HTML]{EFEFEF}0.368 & 0.43 \\
\multirow{-4}{*}{\textbf{Solar}} & 720 & \cellcolor[HTML]{EFEFEF}{\ul 0.244} & {\ul 0.268} & \cellcolor[HTML]{EFEFEF}\textbf{0.223} & \textbf{0.266} & \cellcolor[HTML]{EFEFEF}0.249 & 0.275 & \cellcolor[HTML]{EFEFEF}0.289 & 0.317 & \cellcolor[HTML]{EFEFEF}0.356 & 0.413 & \cellcolor[HTML]{EFEFEF}0.357 & 0.427 & \cellcolor[HTML]{EFEFEF}0.338 & 0.337 & \cellcolor[HTML]{EFEFEF}0.37 & 0.425 \\ \hline
\end{tabular}
\end{table*}

Table~\ref{table_results_all} shows the detailed experimental results on eight open datasets with fixed look-back length $S=96$ and various forecast horizons $T\in \left \{96, 192, 336, 720 \right \}$. MFRS achieves the best performance on almost all datasets, indicating the efficiency and excellent stability in time series forecasting.

Table~\ref{table_seq_ettm1} shows the detailed experimental results on \textbf{ETTm1} with shorter look-back length $S\in \left \{96, 72, 48, 36 \right \}$ and various forecast horizons $T\in \left \{96, 192, 336, 720 \right \}$.

% \usepackage{multirow}
% \usepackage[normalem]{ulem}
\begin{table}[htbp]\small
\setlength{\tabcolsep}{3pt}
\renewcommand{\arraystretch}{1.2}

\caption{Detailed results on \textbf{ETTm1} with shorter look-back length $S\in \left \{96,72,48,36  \right \}$.}
\label{table_seq_ettm1}
\centering
\begin{tabular}{cc|cccccccc}
\hline
\multicolumn{2}{c|}{models} & \multicolumn{2}{c}{MFRS} & \multicolumn{2}{c}{CycleNet} & \multicolumn{2}{c}{PatchTST} & \multicolumn{2}{c}{DLinear} \\
T & S & \textbf{MSE} & \textbf{MAE} & \textbf{MSE} & \textbf{MAE} & \textbf{MSE} & \textbf{MAE} & \textbf{MSE} & \cellcolor[HTML]{FFFFFF}\textbf{MAE} \\ \hline
 & 96 & \textbf{0.314} & \cellcolor[HTML]{FFFFFF}\textbf{0.352} & {\ul 0.328} & \cellcolor[HTML]{FFFFFF}0.372 & 0.329 & \cellcolor[HTML]{FFFFFF}{\ul 0.368} & 0.358 & \cellcolor[HTML]{FFFFFF}0.391 \\
 & 72 & \textbf{0.322} & \cellcolor[HTML]{FFFFFF}\textbf{0.359} & {\ul 0.34} & \cellcolor[HTML]{FFFFFF}{\ul 0.376} & 0.352 & \cellcolor[HTML]{FFFFFF}0.382 & 0.369 & \cellcolor[HTML]{FFFFFF}0.386 \\
 & 48 & \textbf{0.381} & \cellcolor[HTML]{FFFFFF}\textbf{0.388} & {\ul 0.391} & \cellcolor[HTML]{FFFFFF}{\ul 0.396} & 0.457 & \cellcolor[HTML]{FFFFFF}0.42 & 0.499 & \cellcolor[HTML]{FFFFFF}0.447 \\
\multirow{-4}{*}{96} & 36 & \textbf{0.413} & \cellcolor[HTML]{FFFFFF}\textbf{0.404} & {\ul 0.444} & \cellcolor[HTML]{FFFFFF}{\ul 0.423} & 0.589 & \cellcolor[HTML]{FFFFFF}0.47 & 0.583 & \cellcolor[HTML]{FFFFFF}0.49 \\ \hline
 & 96 & \textbf{0.353} & \cellcolor[HTML]{FFFFFF}\textbf{0.375} & 0.38 & \cellcolor[HTML]{FFFFFF}0.399 & {\ul 0.375} & \cellcolor[HTML]{FFFFFF}0.392 & 0.381 & \cellcolor[HTML]{FFFFFF}{\ul 0.391} \\
 & 72 & \textbf{0.367} & \cellcolor[HTML]{FFFFFF}\textbf{0.386} & {\ul 0.381} & \cellcolor[HTML]{FFFFFF}{\ul 0.395} & 0.393 & \cellcolor[HTML]{FFFFFF}0.403 & 0.409 & \cellcolor[HTML]{FFFFFF}0.409 \\
 & 48 & \textbf{0.422} & \cellcolor[HTML]{FFFFFF}\textbf{0.411} & {\ul 0.429} & \cellcolor[HTML]{FFFFFF}{\ul 0.415} & 0.504 & \cellcolor[HTML]{FFFFFF}0.445 & 0.537 & \cellcolor[HTML]{FFFFFF}0.468 \\
\multirow{-4}{*}{192} & 36 & \textbf{0.461} & \cellcolor[HTML]{FFFFFF}\textbf{0.432} & {\ul 0.484} & \cellcolor[HTML]{FFFFFF}{\ul 0.445} & 0.635 & \cellcolor[HTML]{FFFFFF}0.497 & 0.617 & \cellcolor[HTML]{FFFFFF}0.507 \\ \hline
 & 96 & \textbf{0.385} & \cellcolor[HTML]{FFFFFF}\textbf{0.4} & 0.4 & \cellcolor[HTML]{FFFFFF}0.414 & {\ul 0.397} & \cellcolor[HTML]{FFFFFF}{\ul 0.408} & 0.418 & \cellcolor[HTML]{FFFFFF}0.423 \\
 & 72 & \textbf{0.397} & \cellcolor[HTML]{FFFFFF}\textbf{0.408} & {\ul 0.409} & \cellcolor[HTML]{FFFFFF}{\ul 0.414} & 0.422 & \cellcolor[HTML]{FFFFFF}0.421 & 0.444 & \cellcolor[HTML]{FFFFFF}0.435 \\
 & 48 & \textbf{0.465} & \cellcolor[HTML]{FFFFFF}{\ul 0.441} & {\ul 0.47} & \cellcolor[HTML]{FFFFFF}\textbf{0.44} & 0.555 & \cellcolor[HTML]{FFFFFF}0.476 & 0.589 & \cellcolor[HTML]{FFFFFF}0.512 \\
\multirow{-4}{*}{336} & 36 & \textbf{0.496} & \cellcolor[HTML]{FFFFFF}\textbf{0.454} & {\ul 0.516} & \cellcolor[HTML]{FFFFFF}{\ul 0.469} & 0.689 & \cellcolor[HTML]{FFFFFF}0.528 & 0.666 & \cellcolor[HTML]{FFFFFF}0.544 \\ \hline
 & 96 & \textbf{0.443} & \cellcolor[HTML]{FFFFFF}\textbf{0.436} & {\ul 0.46} & \cellcolor[HTML]{FFFFFF}{\ul 0.447} & 0.466 & \cellcolor[HTML]{FFFFFF}0.447 & 0.511 & \cellcolor[HTML]{FFFFFF}0.489 \\
 & 72 & \textbf{0.456} & \cellcolor[HTML]{FFFFFF}\textbf{0.444} & {\ul 0.469} & \cellcolor[HTML]{FFFFFF}{\ul 0.444} & 0.488 & \cellcolor[HTML]{FFFFFF}0.459 & 0.515 & \cellcolor[HTML]{FFFFFF}0.484 \\
 & 48 & \textbf{0.509} & \cellcolor[HTML]{FFFFFF}{\ul 0.466} & {\ul 0.513} & \cellcolor[HTML]{FFFFFF}\textbf{0.465} & 0.601 & \cellcolor[HTML]{FFFFFF}0.501 & 0.635 & \cellcolor[HTML]{FFFFFF}0.545 \\
\multirow{-4}{*}{720} & 36 & \textbf{0.542} & \cellcolor[HTML]{FFFFFF}\textbf{0.482} & {\ul 0.558} & \cellcolor[HTML]{FFFFFF}{\ul 0.493} & 0.717 & \cellcolor[HTML]{FFFFFF}0.548 & 0.698 & \cellcolor[HTML]{FFFFFF}0.573 \\ \hline
\end{tabular}
\end{table}

Table~\ref{table_seq_etth1} shows the detailed experimental results on \textbf{ETTh1} with shorter look-back length $S\in \left \{96, 72, 48, 36 \right \}$ and various forecast horizons $T\in \left \{96, 192, 336, 720 \right \}$.

% \usepackage{multirow}
% \usepackage[normalem]{ulem}
\begin{table}[htbp]\small
\setlength{\tabcolsep}{3pt}
\renewcommand{\arraystretch}{1.2}

\caption{Detailed results on \textbf{ETTh1} with shorter look-back length $S\in \left \{96,72,48,36  \right \}$.}
\label{table_seq_etth1}
\centering
\begin{tabular}{cc|cccccccc}
\hline
\multicolumn{2}{c|}{models} & \multicolumn{2}{c}{MFRS} & \multicolumn{2}{c}{CycleNet} & \multicolumn{2}{c}{PatchTST} & \multicolumn{2}{c}{DLinear} \\
T & S & \textbf{MSE} & \textbf{MAE} & \textbf{MSE} & \textbf{MAE} & \textbf{MSE} & \textbf{MAE} & \textbf{MSE} & \textbf{MAE} \\ \hline
 & 96 & \textbf{0.371} & \cellcolor[HTML]{FFFFFF}\textbf{0.395} & {\ul 0.381} & \cellcolor[HTML]{FFFFFF}0.404 & 0.392 & \cellcolor[HTML]{FFFFFF}{\ul 0.403} & 0.42 & \cellcolor[HTML]{FFFFFF}0.429 \\
 & 72 & \textbf{0.371} & \cellcolor[HTML]{FFFFFF}\textbf{0.395} & {\ul 0.378} & \cellcolor[HTML]{FFFFFF}0.402 & 0.391 & \cellcolor[HTML]{FFFFFF}{\ul 0.4} & 0.424 & \cellcolor[HTML]{FFFFFF}0.435 \\
 & 48 & \textbf{0.374} & \cellcolor[HTML]{FFFFFF}\textbf{0.396} & {\ul 0.38} & \cellcolor[HTML]{FFFFFF}0.402 & 0.393 & \cellcolor[HTML]{FFFFFF}{\ul 0.4} & 0.404 & \cellcolor[HTML]{FFFFFF}0.414 \\
\multirow{-4}{*}{96} & 36 & \textbf{0.387} & \cellcolor[HTML]{FFFFFF}\textbf{0.402} & {\ul 0.391} & \cellcolor[HTML]{FFFFFF}{\ul 0.408} & 0.414 & \cellcolor[HTML]{FFFFFF}0.411 & 0.418 & \cellcolor[HTML]{FFFFFF}0.422 \\ \hline
 & 96 & \textbf{0.422} & \cellcolor[HTML]{FFFFFF}\textbf{0.425} & {\ul 0.443} & \cellcolor[HTML]{FFFFFF}{\ul 0.43} & 0.453 & \cellcolor[HTML]{FFFFFF}0.435 & 0.448 & \cellcolor[HTML]{FFFFFF}0.443 \\
 & 72 & \textbf{0.424} & \cellcolor[HTML]{FFFFFF}\textbf{0.426} & {\ul 0.428} & \cellcolor[HTML]{FFFFFF}0.432 & 0.447 & \cellcolor[HTML]{FFFFFF}{\ul 0.431} & 0.498 & \cellcolor[HTML]{FFFFFF}0.486 \\
 & 48 & \textbf{0.431} & \cellcolor[HTML]{FFFFFF}\textbf{0.429} & {\ul 0.437} & \cellcolor[HTML]{FFFFFF}{\ul 0.432} & 0.45 & \cellcolor[HTML]{FFFFFF}0.433 & 0.451 & \cellcolor[HTML]{FFFFFF}0.441 \\
\multirow{-4}{*}{192} & 36 & {\ul 0.445} & \cellcolor[HTML]{FFFFFF}\textbf{0.435} & \textbf{0.442} & \cellcolor[HTML]{FFFFFF}{\ul 0.44} & 0.469 & \cellcolor[HTML]{FFFFFF}0.442 & 0.471 & \cellcolor[HTML]{FFFFFF}0.456 \\ \hline
 & 96 & \textbf{0.465} & \cellcolor[HTML]{FFFFFF}\textbf{0.445} & {\ul 0.483} & \cellcolor[HTML]{FFFFFF}{\ul 0.449} & 0.49 & \cellcolor[HTML]{FFFFFF}0.452 & 0.483 & \cellcolor[HTML]{FFFFFF}0.459 \\
 & 72 & \textbf{0.471} & \cellcolor[HTML]{FFFFFF}\textbf{0.45} & {\ul 0.489} & \cellcolor[HTML]{FFFFFF}{\ul 0.451} & 0.501 & \cellcolor[HTML]{FFFFFF}0.459 & 0.488 & \cellcolor[HTML]{FFFFFF}0.462 \\
 & 48 & \textbf{0.478} & \cellcolor[HTML]{FFFFFF}\textbf{0.454} & {\ul 0.478} & \cellcolor[HTML]{FFFFFF}{\ul 0.454} & 0.508 & \cellcolor[HTML]{FFFFFF}0.464 & 0.504 & \cellcolor[HTML]{FFFFFF}0.473 \\
\multirow{-4}{*}{336} & 36 & \textbf{0.492} & \cellcolor[HTML]{FFFFFF}\textbf{0.46} & {\ul 0.501} & \cellcolor[HTML]{FFFFFF}{\ul 0.469} & 0.527 & \cellcolor[HTML]{FFFFFF}0.472 & 0.516 & \cellcolor[HTML]{FFFFFF}0.48 \\ \hline
 & 96 & \textbf{0.469} & \cellcolor[HTML]{FFFFFF}\textbf{0.466} & {\ul 0.491} & \cellcolor[HTML]{FFFFFF}{\ul 0.471} & 0.535 & \cellcolor[HTML]{FFFFFF}0.498 & 0.571 & \cellcolor[HTML]{FFFFFF}0.545 \\
 & 72 & \textbf{0.489} & \cellcolor[HTML]{FFFFFF}\textbf{0.477} & {\ul 0.493} & \cellcolor[HTML]{FFFFFF}{\ul 0.484} & 0.521 & \cellcolor[HTML]{FFFFFF}0.488 & 0.546 & \cellcolor[HTML]{FFFFFF}0.529 \\
 & 48 & \textbf{0.493} & \cellcolor[HTML]{FFFFFF}\textbf{0.479} & 0.54 & \cellcolor[HTML]{FFFFFF}0.506 & {\ul 0.51} & \cellcolor[HTML]{FFFFFF}{\ul 0.481} & 0.542 & \cellcolor[HTML]{FFFFFF}0.524 \\
\multirow{-4}{*}{720} & 36 & \textbf{0.504} & \cellcolor[HTML]{FFFFFF}\textbf{0.483} & {\ul 0.532} & \cellcolor[HTML]{FFFFFF}{\ul 0.5} & 0.54 & \cellcolor[HTML]{FFFFFF}0.5 & 0.548 & \cellcolor[HTML]{FFFFFF}0.526 \\ \hline
\end{tabular}
\end{table}

\section{Synthetic Datasets}

\subsection{Synthetic Method}
\label{app2-syn-method}

We propose some self-created synthetic datasets $\textbf{Compose}$ as simple yet reliable benchmarks for time series forcasting. $\textbf{Compose}$ consists of two components: deterministic signal $\textbf{Z}$ with definite values at any time point and random signal $\textbf{U}$ following a certain distribution. Specifically, the $i$-th channel $\mathbf{Z }^{\left ( i \right ) }$ is synthesized by 4 sine with period $\mathcal{T} \in\left \{ 72, 36, 24, 18 \right \}$ or $\left \{ 720, 360, 240, 180 \right \}$, expressed as $\mathbf{Z} ^{\left ( i \right ) } \left ( t \right ) =\sum_{k=1}^{4} \textrm{A}_{k}^{\left ( i \right ) }\ast sin\left ( \frac{2\pi t}{\mathcal{T}_{k} }  \right )$, where the amplitudes $\textrm{A}^{\left ( i \right ) }$ are set different values across channels. Let $\mathbf{Z}_{1}$ and $\mathbf{Z}_{2}$ denote the series with shorter and longer period respectively. The random signal has two types: $\mathbf{U} _{1} \sim N\left ( \mu,\sigma ^{2}  \right )$ and $\mathbf{U} _{2} \sim Possion\left ( \lambda  \right )$. $\mathbf{U} _{1}$ is first generated as a standard normally distributed signal using function np.random.randn(), and then converted into a Gaussian signal with mean $\mu$ and variance $\sigma$ by formula $\mathbf{U}_{1} = \sigma \ast \mathbf{U}_{1}+\mu$. $\mathbf{U} _{2}$ is generated as a Poisson signal with intensity $\lambda$ using function np.random.poisson( $\lambda$ ).

By simply adding $\mathbf{U}$ and $\mathbf{Z}$ together, we obtain the four sets of synthetic datasets: $\textbf{Compose}_{1} =\mathbf{Z}_{1}+  {\mathbf{U}}_{1}$, $\textbf{Compose}_{2} =\mathbf{Z}_{2}+  {\mathbf{U}}_{1}$, $\textbf{Compose}_{3} =\mathbf{Z}_{1}+  {\mathbf{U}}_{2}$, $\textbf{Compose}_{4} =\mathbf{\mathbf{U}}_{2}+  {\mathbf{U}}_{2}$.

\subsection{Random Singal}
\label{app2-random}

\begin{figure}[h]
\centering
\subfigure[$\mathbf{U}_{1}  \sim N\left ( -2,1 \right )$]
{
    \begin{minipage}[b]{\linewidth}
        \centering
        \includegraphics[width=0.48\columnwidth]{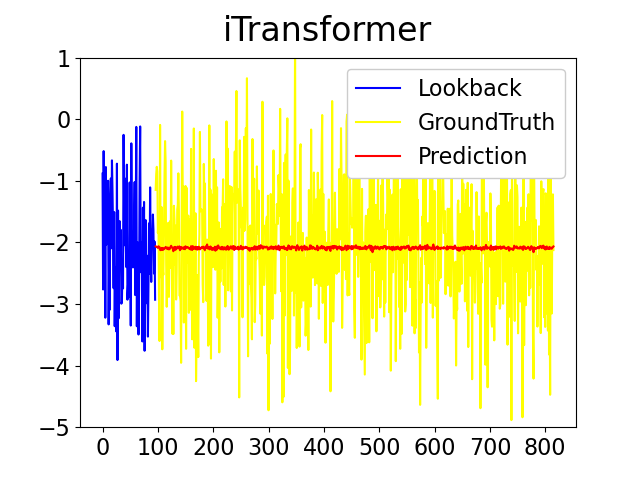}
        \includegraphics[width=0.48\columnwidth]{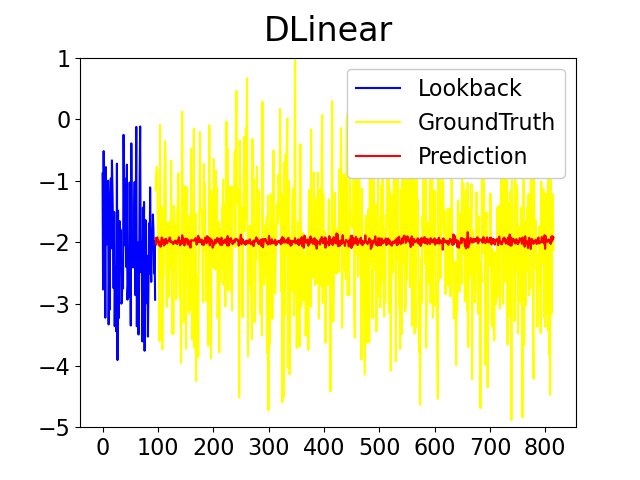}
    \end{minipage}
}
\subfigure[$\mathbf{U}_{1}  \sim N\left ( 2,1 \right )$]
{
    \begin{minipage}[b]{\linewidth}
        \centering
        \includegraphics[width=0.48\columnwidth]{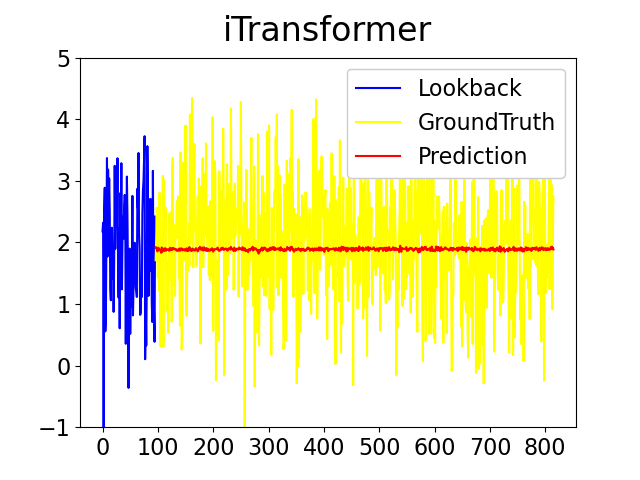}
        \includegraphics[width=0.48\columnwidth]{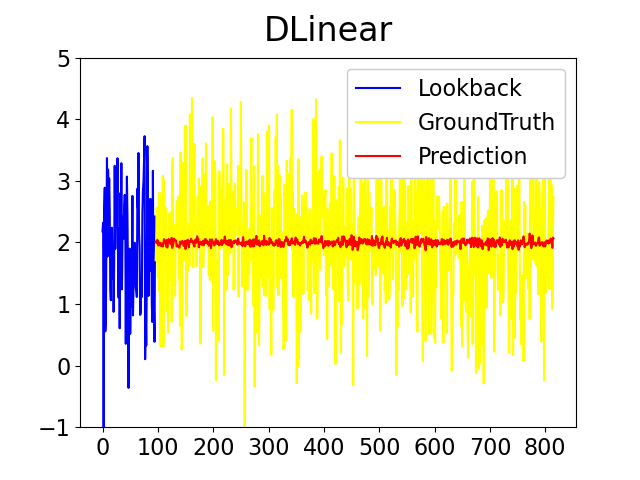}
    \end{minipage}
}
\subfigure[$\mathbf{U}_{2}  \sim Possion\left ( 1 \right )$]
{
    \begin{minipage}[b]{\linewidth}
        \centering
        \includegraphics[width=0.48\columnwidth]{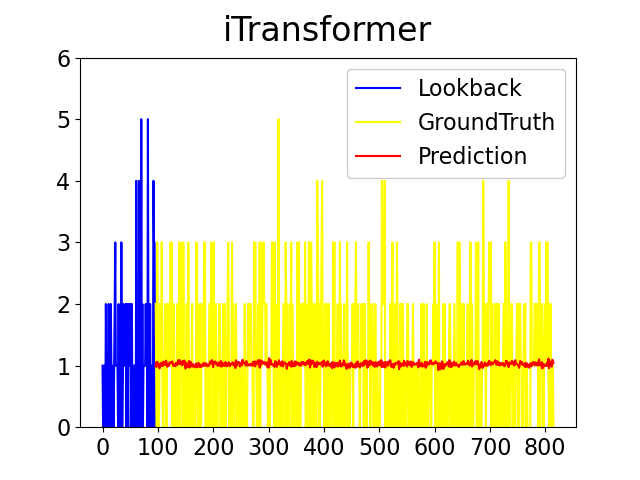}
        \includegraphics[width=0.48\columnwidth]{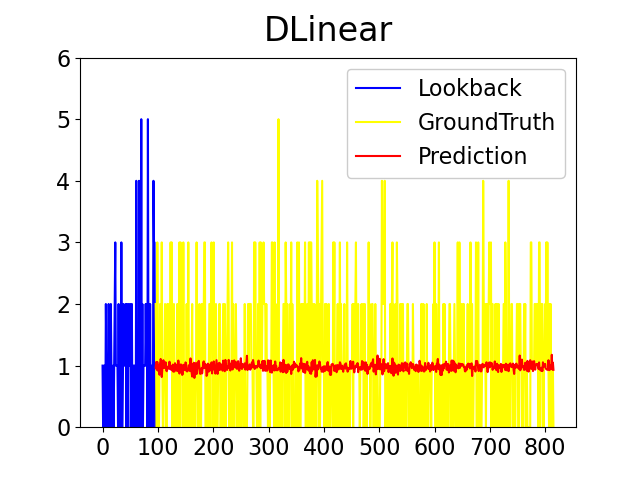}
    \end{minipage}
}
\subfigure[$\mathbf{U}_{2}  \sim Possion\left ( 3 \right )$]
{
    \begin{minipage}[b]{\linewidth}
        \centering
        \includegraphics[width=0.48\columnwidth]{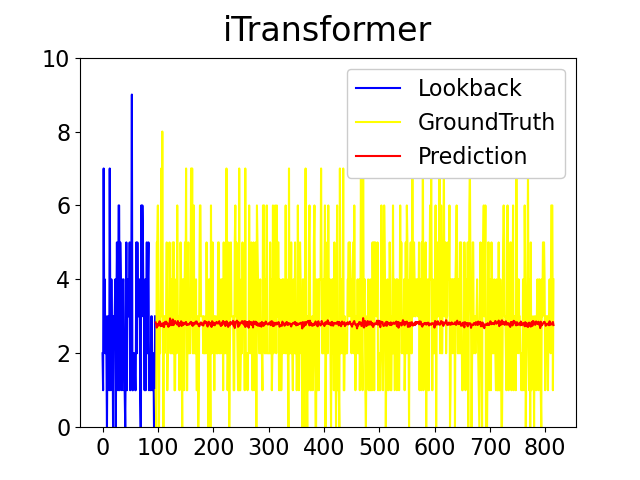}
        \includegraphics[width=0.48\columnwidth]{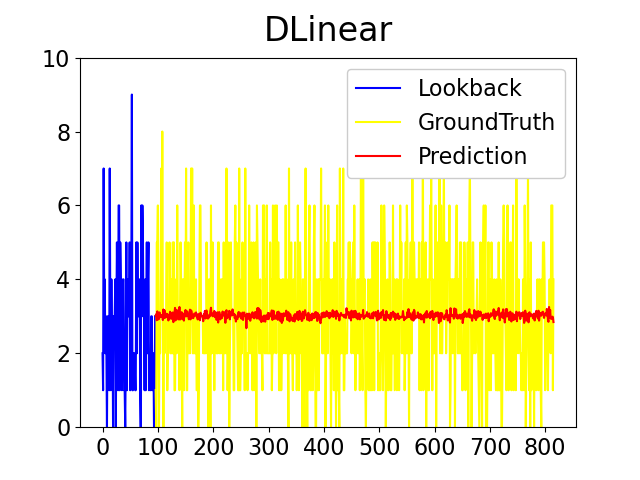}
    \end{minipage}
}

\vskip -0.1in
\caption{Visual results on random signal}
\label{figure-mean1}
\end{figure}

The loss of random signal can be calculated using the following formula:

\begin{equation*}
    \begin{split}
        \mathbf{E} \left ( \textrm{MSELoss} \right ) &=\mathbf{E} \left ( \sum_{t}^{} \left ( \hat{{\mathbf{U}} }-{\mathbf{U}}   \right )^{2}   \right ) \\
        &=\sum_{t}^{} \left \{ \left ( \hat{{\mathbf{U}} }-\mathbf{E} \left ( {\mathbf{U}}  \right )   \right )^{2}+\mathbf{D} \left ( {\mathbf{U}}  \right )  \right \}
    \end{split}
\end{equation*}

Where $\mathbf{D} \left ( \cdot  \right )$ denotes the variance of random variable. It can be concluded that, the expectation of a random signal minimizes its training MSELoss, thus the optimal prediction $\hat{\mathbf{U}}=\mathbf{E} \left ( \mathbf{U}  \right )$. Then we validate this inference by visualizing the random signal prediction results, as shown in Figure~\ref{figure-mean1}.

The left side in Figure~\ref{figure-mean1} show the prediction results modeled on iTransformer, while the right side on DLinear. It can be seen that ${\mathbf{U}} _{1}$ only fluctuate slightly around $\mathbf{E} \left ({\mathbf{U}} _{1} \right )=\mu$, and ${\mathbf{U}} _{2}$ around $\mathbf{E} \left ( {\mathbf{U}} _{2} \right )=\lambda$, which aligns with our initial judgement. Moreover, the jitter amplitude of iTransformer is smaller compared to that of DLinear for both ${\mathbf{U}} _{1}$ and ${\mathbf{U}} _{2}$.

\subsection{Optimal Forecaster}
\label{app2-optimal}

Based on above claims, we define the theoretically optimal forecaster $f:\hat{\mathbf{Z} }_{t}=\mathbf{Z}_{t},\hat{{\mathbf{U}} }=\mathbf{E} \left ({\mathbf{U}}  \right )$. Subsequently, the optimal evaluation metrics of $f$ are derived through the following formulas.

\vskip -0.1in

\begin{equation*}
    \begin{split}
        \textrm{MSE}_{\textrm{Optimal}} &=\mathbf{E} \left ( \frac{1}{T} \sum_{t=1}^{T} \left ( \hat{\mathbf{Z} }_{t} +\hat{\mathbf{U} }-\mathbf{Z}_{t} -\mathbf{U} \right )^{2} \right )  \\
        &= \mathbf{D} \left (\mathbf{U}  \right )
    \end{split}
\end{equation*}

\begin{equation*}
    \textrm{MAE}_{\textrm{Optimal}} = \mathbf{E}  \left ( \left | \mathbf{U} -\mathbf{E} \left ( \mathbf{U}  \right )  \right | \right )    
\end{equation*}

For $\textbf{Compose}_{1}$ and $\textbf{Compose}_{2}$, we have

\begin{equation*}
    \textrm{MSE}_{\textrm{Optimal}} =\sigma ^{2}    
\end{equation*}

\vskip -0.15in

\begin{equation*}
    \begin{split}
        \textrm{MAE}_{\textrm{Optimal}} &= \int_{-\infty }^{\infty } \left | x \right | \ast \frac{1}{\sqrt{2\pi \sigma } }e^{-\frac{x^{2} }{2\delta ^{2} } }\\
        &=\sqrt{\frac{2}{\pi } }\sigma
    \end{split}
\end{equation*}

For $\textbf{Compose}_{3}$ and $\textbf{Compose}_{4}$, we have

\begin{equation*}
    \textrm{MSE}_{\textrm{Optimal}} =\lambda    
\end{equation*}

\vskip -0.15in

\begin{equation*}
    \begin{split}
        \textrm{MAE}_{\textrm{Optimal}} &=\sum_{k=0}^{\infty } \left | k-\lambda  \right |\frac{\lambda ^{k} }{k!}e^{-\lambda }\\
        &=2e^{-\lambda } \sum_{k=0}^{\lambda -1}\left ( \lambda -k \right )\frac{\lambda ^{k} }{k!}
    \end{split}
\end{equation*}

\subsection{Results Analysis}
\label{app2-results}

By varying the parameter $\sigma\in \left ( 0,1,2,3,4,5 \right )$ of the random signal $\mathbf{U}_{1}$ with fixed $\mu =0$, we can obtain a series of different $\textbf{Compose}_{1}$ and $\textbf{Compose}_{2}$. Similarly, a series of different $\textbf{Compose}_{3}$ and $\textbf{Compose}_{4}$ can be obtained by varying the parameter $\lambda\in \left ( 1,2,3,4,5 \right )$ of $\mathbf{U}_{2}$. The four groups of synthetic datasets, totaling 22, are then experimented with MFRS, iTransformer, and DLinear. We provide all results in detail alongside the theoretical optimal ones in Table~\ref{table-compose1}~\ref{table-compose2}~\ref{table-compose3}~\ref{table-compose4}.

%\usepackage{multirow}
\begin{table*}[h]
\setlength{\tabcolsep}{8pt}
\renewcommand{\arraystretch}{1.5}
\caption{Results on $\textbf{Compose}_{1}$}
\label{table-compose1}
\vskip 0.15in
\centering
\begin{tabular}{|cc|cc|cc|cc|cc|}
\hline
\multicolumn{2}{|c|}{models} & \multicolumn{2}{c|}{\textbf{MFRS}} & \multicolumn{2}{c|}{\textbf{iTransformer}} & \multicolumn{2}{c|}{\textbf{DLinear}} & \multicolumn{2}{c|}{\textbf{Optimal}} \\ \hline
\multicolumn{1}{|c|}{$\sigma$} & T & \multicolumn{1}{c|}{\textbf{MSE}} & \textbf{MAE} & \multicolumn{1}{c|}{\textbf{MSE}} & \textbf{MAE} & \multicolumn{1}{c|}{\textbf{MSE}} & \textbf{MAE} & \multicolumn{1}{c|}{\textbf{MSE}} & \textbf{MAE} \\ \hline
\multicolumn{1}{|c|}{\multirow{2}{*}{0}} & 96 & \multicolumn{1}{c|}{0} & 0.013 & \multicolumn{1}{c|}{0} & 0.012 & \multicolumn{1}{c|}{0} & 0 & \multicolumn{1}{c|}{\multirow{2}{*}{0}} & \multirow{2}{*}{0} \\ \cline{2-8}
\multicolumn{1}{|c|}{} & 720 & \multicolumn{1}{c|}{0} & 0.012 & \multicolumn{1}{c|}{0} & 0.011 & \multicolumn{1}{c|}{0} & 0 & \multicolumn{1}{c|}{} &  \\ \hline
\multicolumn{1}{|c|}{\multirow{2}{*}{1}} & 96 & \multicolumn{1}{c|}{1.037} & 0.813 & \multicolumn{1}{c|}{1.042} & 0.814 & \multicolumn{1}{c|}{1.095} & 0.835 & \multicolumn{1}{c|}{\multirow{2}{*}{1}} & \multirow{2}{*}{0.798} \\ \cline{2-8}
\multicolumn{1}{|c|}{} & 720 & \multicolumn{1}{c|}{1.038} & 0.814 & \multicolumn{1}{c|}{1.042} & 0.815 & \multicolumn{1}{c|}{1.096} & 0.836 & \multicolumn{1}{c|}{} &  \\ \hline
\multicolumn{1}{|c|}{\multirow{2}{*}{2}} & 96 & \multicolumn{1}{c|}{4.17} & 1.63 & \multicolumn{1}{c|}{4.188} & 1.633 & \multicolumn{1}{c|}{4.359} & 1.665 & \multicolumn{1}{c|}{\multirow{2}{*}{4}} & \multirow{2}{*}{1.596} \\ \cline{2-8}
\multicolumn{1}{|c|}{} & 720 & \multicolumn{1}{c|}{4.111} & 1.618 & \multicolumn{1}{c|}{4.12} & 1.62 & \multicolumn{1}{c|}{4.324} & 1.658 & \multicolumn{1}{c|}{} &  \\ \hline
\multicolumn{1}{|c|}{\multirow{2}{*}{3}} & 96 & \multicolumn{1}{c|}{9.321} & 2.435 & \multicolumn{1}{c|}{9.356} & 2.44 & \multicolumn{1}{c|}{9.656} & 2.48 & \multicolumn{1}{c|}{\multirow{2}{*}{9}} & \multirow{2}{*}{2.394} \\ \cline{2-8}
\multicolumn{1}{|c|}{} & 720 & \multicolumn{1}{c|}{9.232} & 2.422 & \multicolumn{1}{c|}{9.266} & 2.426 & \multicolumn{1}{c|}{9.664} & 2.479 & \multicolumn{1}{c|}{} &  \\ \hline
\multicolumn{1}{|c|}{\multirow{2}{*}{4}} & 96 & \multicolumn{1}{c|}{16.362} & 3.237 & \multicolumn{1}{c|}{16.342} & 3.234 & \multicolumn{1}{c|}{16.982} & 3.3 & \multicolumn{1}{c|}{\multirow{2}{*}{16}} & \multirow{2}{*}{3.192} \\ \cline{2-8}
\multicolumn{1}{|c|}{} & 720 & \multicolumn{1}{c|}{16.257} & 3.223 & \multicolumn{1}{c|}{16.296} & 3.227 & \multicolumn{1}{c|}{16.942} & 3.289 & \multicolumn{1}{c|}{} &  \\ \hline
\multicolumn{1}{|c|}{\multirow{2}{*}{5}} & 96 & \multicolumn{1}{c|}{26.172} & 4.097 & \multicolumn{1}{c|}{26.119} & 4.093 & \multicolumn{1}{c|}{26.894} & 4.151 & \multicolumn{1}{c|}{\multirow{2}{*}{25}} & \multirow{2}{*}{3.989} \\ \cline{2-8}
\multicolumn{1}{|c|}{} & 720 & \multicolumn{1}{c|}{26.271} & 4.107 & \multicolumn{1}{c|}{26.329} & 4.113 & \multicolumn{1}{c|}{27.019} & 4.164 & \multicolumn{1}{c|}{} &  \\ \hline
\end{tabular}
\end{table*}

% \usepackage{multirow}
\begin{table*}[t]
\setlength{\tabcolsep}{8pt}
\renewcommand{\arraystretch}{1.5}
\caption{Results on $\textbf{Compose}_{2}$}
\label{table-compose2}
\vskip 0.15in
\centering
\begin{tabular}{|cc|cc|cc|cc|cc|}
\hline
\multicolumn{2}{|c|}{models} & \multicolumn{2}{c|}{\textbf{MFRS}} & \multicolumn{2}{c|}{\textbf{iTransformer}} & \multicolumn{2}{c|}{\textbf{DLinear}} & \multicolumn{2}{c|}{\textbf{Optimal}} \\ \hline
\multicolumn{1}{|c|}{$\sigma$} & T & \multicolumn{1}{c|}{\textbf{MSE}} & \textbf{MAE} & \multicolumn{1}{c|}{\textbf{MSE}} & \textbf{MAE} & \multicolumn{1}{c|}{\textbf{MSE}} & \textbf{MAE} & \multicolumn{1}{c|}{\textbf{MSE}} & \textbf{MAE} \\ \hline
\multicolumn{1}{|c|}{\multirow{2}{*}{0}} & 96 & \multicolumn{1}{c|}{0.01} & 0.064 & \multicolumn{1}{c|}{0.043} & 0.135 & \multicolumn{1}{c|}{0.615} & 0.512 & \multicolumn{1}{c|}{\multirow{2}{*}{0}} & \multirow{2}{*}{0} \\ \cline{2-8}
\multicolumn{1}{|c|}{} & 720 & \multicolumn{1}{c|}{0.022} & 0.084 & \multicolumn{1}{c|}{2.472} & 1.09 & \multicolumn{1}{c|}{1.838} & 0.913 & \multicolumn{1}{c|}{} &  \\ \hline
\multicolumn{1}{|c|}{\multirow{2}{*}{1}} & 96 & \multicolumn{1}{c|}{1.086} & 0.83 & \multicolumn{1}{c|}{1.431} & 0.94 & \multicolumn{1}{c|}{2.014} & 1.11 & \multicolumn{1}{c|}{\multirow{2}{*}{1}} & \multirow{2}{*}{0.798} \\ \cline{2-8}
\multicolumn{1}{|c|}{} & 720 & \multicolumn{1}{c|}{1.113} & 0.839 & \multicolumn{1}{c|}{4.352} & 1.582 & \multicolumn{1}{c|}{3.079} & 1.354 & \multicolumn{1}{c|}{} &  \\ \hline
\multicolumn{1}{|c|}{\multirow{2}{*}{2}} & 96 & \multicolumn{1}{c|}{4.099} & 1.615 & \multicolumn{1}{c|}{4.682} & 1.72 & \multicolumn{1}{c|}{5.257} & 1.822 & \multicolumn{1}{c|}{\multirow{2}{*}{4}} & \multirow{2}{*}{1.596} \\ \cline{2-8}
\multicolumn{1}{|c|}{} & 720 & \multicolumn{1}{c|}{4.095} & 1.616 & \multicolumn{1}{c|}{6.438} & 1.986 & \multicolumn{1}{c|}{6.282} & 1.99 & \multicolumn{1}{c|}{} &  \\ \hline
\multicolumn{1}{|c|}{\multirow{2}{*}{3}} & 96 & \multicolumn{1}{c|}{9.262} & 2.426 & \multicolumn{1}{c|}{10.527} & 2.573 & \multicolumn{1}{c|}{10.551} & 2.582 & \multicolumn{1}{c|}{\multirow{2}{*}{9}} & \multirow{2}{*}{2.394} \\ \cline{2-8}
\multicolumn{1}{|c|}{} & 720 & \multicolumn{1}{c|}{9.189} & 2.415 & \multicolumn{1}{c|}{12.471} & 2.802 & \multicolumn{1}{c|}{11.36} & 2.679 & \multicolumn{1}{c|}{} &  \\ \hline
\multicolumn{1}{|c|}{\multirow{2}{*}{4}} & 96 & \multicolumn{1}{c|}{16.369} & 3.227 & \multicolumn{1}{c|}{17.686} & 3.354 & \multicolumn{1}{c|}{17.702} & 3.36 & \multicolumn{1}{c|}{\multirow{2}{*}{16}} & \multirow{2}{*}{3.192} \\ \cline{2-8}
\multicolumn{1}{|c|}{} & 720 & \multicolumn{1}{c|}{16.121} & 3.199 & \multicolumn{1}{c|}{19.889} & 3.559 & \multicolumn{1}{c|}{18.29} & 3.413 & \multicolumn{1}{c|}{} &  \\ \hline
\multicolumn{1}{|c|}{\multirow{2}{*}{5}} & 96 & \multicolumn{1}{c|}{25.83} & 4.057 & \multicolumn{1}{c|}{27.302} & 4.157 & \multicolumn{1}{c|}{27} & 4.141 & \multicolumn{1}{c|}{\multirow{2}{*}{25}} & \multirow{2}{*}{3.989} \\ \cline{2-8}
\multicolumn{1}{|c|}{} & 720 & \multicolumn{1}{c|}{25.62} & 4.039 & \multicolumn{1}{c|}{28.982} & 4.291 & \multicolumn{1}{c|}{27.591} & 4.187 & \multicolumn{1}{c|}{} &  \\ \hline
\end{tabular}
\end{table*}

% \usepackage{multirow}
\begin{table*}[t]
\setlength{\tabcolsep}{8pt}
\renewcommand{\arraystretch}{1.5}
\caption{Results on $\textbf{Compose}_{3}$}
\label{table-compose3}
\vskip 0.15in
\centering
\begin{tabular}{|cc|cc|cc|cc|cc|}
\hline
\multicolumn{2}{|c|}{models} & \multicolumn{2}{c|}{\textbf{MFRS}} & \multicolumn{2}{c|}{\textbf{iTransformer}} & \multicolumn{2}{c|}{\textbf{DLinear}} & \multicolumn{2}{c|}{\textbf{Optimal}} \\ \hline
\multicolumn{1}{|c|}{$\lambda$} & T & \multicolumn{1}{c|}{\textbf{MSE}} & \textbf{MAE} & \multicolumn{1}{c|}{\textbf{MSE}} & \textbf{MAE} & \multicolumn{1}{c|}{\textbf{MSE}} & \textbf{MAE} & \multicolumn{1}{c|}{\textbf{MSE}} & \textbf{MAE} \\ \hline
\multicolumn{1}{|c|}{\multirow{2}{*}{1}} & 96 & \multicolumn{1}{c|}{1.035} & 0.788 & \multicolumn{1}{c|}{1.039} & 0.792 & \multicolumn{1}{c|}{1.094} & 0.825 & \multicolumn{1}{c|}{\multirow{2}{*}{1}} & \multirow{2}{*}{0.736} \\ \cline{2-8}
\multicolumn{1}{|c|}{} & 720 & \multicolumn{1}{c|}{1.031} & 0.784 & \multicolumn{1}{c|}{1.035} & 0.787 & \multicolumn{1}{c|}{1.093} & 0.825 & \multicolumn{1}{c|}{} &  \\ \hline
\multicolumn{1}{|c|}{\multirow{2}{*}{2}} & 96 & \multicolumn{1}{c|}{2.06} & 1.134 & \multicolumn{1}{c|}{2.071} & 1.138 & \multicolumn{1}{c|}{2.18} & 1.177 & \multicolumn{1}{c|}{\multirow{2}{*}{2}} & \multirow{2}{*}{1.083} \\ \cline{2-8}
\multicolumn{1}{|c|}{} & 720 & \multicolumn{1}{c|}{2.061} & 1.134 & \multicolumn{1}{c|}{2.066} & 1.136 & \multicolumn{1}{c|}{2.187} & 1.179 & \multicolumn{1}{c|}{} &  \\ \hline
\multicolumn{1}{|c|}{\multirow{2}{*}{3}} & 96 & \multicolumn{1}{c|}{3.056} & 1.384 & \multicolumn{1}{c|}{3.073} & 1.389 & \multicolumn{1}{c|}{3.208} & 1.423 & \multicolumn{1}{c|}{\multirow{2}{*}{3}} & \multirow{2}{*}{1.344} \\ \cline{2-8}
\multicolumn{1}{|c|}{} & 720 & \multicolumn{1}{c|}{3.042} & 1.38 & \multicolumn{1}{c|}{3.051} & 1.383 & \multicolumn{1}{c|}{3.201} & 1.421 & \multicolumn{1}{c|}{} &  \\ \hline
\multicolumn{1}{|c|}{\multirow{2}{*}{4}} & 96 & \multicolumn{1}{c|}{4.055} & 1.595 & \multicolumn{1}{c|}{4.071} & 1.6 & \multicolumn{1}{c|}{4.286} & 1.648 & \multicolumn{1}{c|}{\multirow{2}{*}{4}} & \multirow{2}{*}{1.563} \\ \cline{2-8}
\multicolumn{1}{|c|}{} & 720 & \multicolumn{1}{c|}{4.048} & 1.593 & \multicolumn{1}{c|}{4.056} & 1.595 & \multicolumn{1}{c|}{4.287} & 1.647 & \multicolumn{1}{c|}{} &  \\ \hline
\multicolumn{1}{|c|}{\multirow{2}{*}{5}} & 96 & \multicolumn{1}{c|}{4.976} & 1.777 & \multicolumn{1}{c|}{4.997} & 1.781 & \multicolumn{1}{c|}{5.226} & 1.827 & \multicolumn{1}{c|}{\multirow{2}{*}{5}} & \multirow{2}{*}{1.755} \\ \cline{2-8}
\multicolumn{1}{|c|}{} & 720 & \multicolumn{1}{c|}{4.944} & 1.767 & \multicolumn{1}{c|}{4.958} & 1.77 & \multicolumn{1}{c|}{5.207} & 1.818 & \multicolumn{1}{c|}{} &  \\ \hline
\end{tabular}
\end{table*}

% \usepackage{multirow}
\begin{table*}[t]
\setlength{\tabcolsep}{8pt}
\renewcommand{\arraystretch}{1.5}
\caption{Results on $\textbf{Compose}_{4}$}
\label{table-compose4}
\vskip 0.15in
\centering
\begin{tabular}{|cc|cc|cc|cc|cc|}
\hline
\multicolumn{2}{|c|}{models} & \multicolumn{2}{c|}{\textbf{MFRS}} & \multicolumn{2}{c|}{\textbf{iTransformer}} & \multicolumn{2}{c|}{\textbf{DLinear}} & \multicolumn{2}{c|}{\textbf{Optimal}} \\ \hline
\multicolumn{1}{|c|}{$\lambda$} & T & \multicolumn{1}{c|}{\textbf{MSE}} & \textbf{MAE} & \multicolumn{1}{c|}{\textbf{MSE}} & \textbf{MAE} & \multicolumn{1}{c|}{\textbf{MSE}} & \textbf{MAE} & \multicolumn{1}{c|}{\textbf{MSE}} & \textbf{MAE} \\ \hline
\multicolumn{1}{|c|}{\multirow{2}{*}{1}} & 96 & \multicolumn{1}{c|}{1.086} & 0.815 & \multicolumn{1}{c|}{1.537} & 0.955 & \multicolumn{1}{c|}{2.043} & 1.111 & \multicolumn{1}{c|}{\multirow{2}{*}{1}} & \multirow{2}{*}{0.736} \\ \cline{2-8}
\multicolumn{1}{|c|}{} & 720 & \multicolumn{1}{c|}{1.122} & 0.831 & \multicolumn{1}{c|}{3.774} & 1.503 & \multicolumn{1}{c|}{3.055} & 1.343 & \multicolumn{1}{c|}{} &  \\ \hline
\multicolumn{1}{|c|}{\multirow{2}{*}{2}} & 96 & \multicolumn{1}{c|}{2.124} & 1.156 & \multicolumn{1}{c|}{2.757} & 1.301 & \multicolumn{1}{c|}{3.165} & 1.403 & \multicolumn{1}{c|}{\multirow{2}{*}{2}} & \multirow{2}{*}{1.083} \\ \cline{2-8}
\multicolumn{1}{|c|}{} & 720 & \multicolumn{1}{c|}{2.183} & 1.173 & \multicolumn{1}{c|}{4.237} & 1.565 & \multicolumn{1}{c|}{4.222} & 1.61 & \multicolumn{1}{c|}{} &  \\ \hline
\multicolumn{1}{|c|}{\multirow{2}{*}{3}} & 96 & \multicolumn{1}{c|}{3.18} & 1.417 & \multicolumn{1}{c|}{3.897} & 1.559 & \multicolumn{1}{c|}{4.292} & 1.638 & \multicolumn{1}{c|}{\multirow{2}{*}{3}} & \multirow{2}{*}{1.344} \\ \cline{2-8}
\multicolumn{1}{|c|}{} & 720 & \multicolumn{1}{c|}{3.221} & 1.422 & \multicolumn{1}{c|}{5.794} & 1.89 & \multicolumn{1}{c|}{5.368} & 1.825 & \multicolumn{1}{c|}{} &  \\ \hline
\multicolumn{1}{|c|}{\multirow{2}{*}{4}} & 96 & \multicolumn{1}{c|}{4.219} & 1.635 & \multicolumn{1}{c|}{5.014} & 1.766 & \multicolumn{1}{c|}{5.463} & 1.853 & \multicolumn{1}{c|}{\multirow{2}{*}{4}} & \multirow{2}{*}{1.563} \\ \cline{2-8}
\multicolumn{1}{|c|}{} & 720 & \multicolumn{1}{c|}{4.25} & 1.641 & \multicolumn{1}{c|}{7.406} & 2.121 & \multicolumn{1}{c|}{6.407} & 2.01 & \multicolumn{1}{c|}{} &  \\ \hline
\multicolumn{1}{|c|}{\multirow{2}{*}{5}} & 96 & \multicolumn{1}{c|}{5.121} & 1.807 & \multicolumn{1}{c|}{5.905} & 1.931 & \multicolumn{1}{c|}{6.349} & 2.006 & \multicolumn{1}{c|}{\multirow{2}{*}{5}} & \multirow{2}{*}{1.755} \\ \cline{2-8}
\multicolumn{1}{|c|}{} & 720 & \multicolumn{1}{c|}{5.113} & 1.809 & \multicolumn{1}{c|}{8.203} & 2.278 & \multicolumn{1}{c|}{7.26} & 2.15 & \multicolumn{1}{c|}{} &  \\ \hline
\end{tabular}
\end{table*}

%%%%%%%%%%%%%%%%%%%%%%%%%%%%%%%%%%%%%%%%%%%%%%%%%%%%%%%%%%%%%%%%%%%%%%%%%%%%%%%
%%%%%%%%%%%%%%%%%%%%%%%%%%%%%%%%%%%%%%%%%%%%%%%%%%%%%%%%%%%%%%%%%%%%%%%%%%%%%%%

\end{document}